\documentclass[lettersize,journal]{IEEEtran}
\usepackage{amsmath,amsfonts}
\usepackage{algorithmic}
\usepackage{algorithm}
\usepackage{array}
\usepackage[caption=false,font=normalsize,labelfont=sf,textfont=sf]{subfig}
\usepackage{textcomp}
\usepackage{stfloats}
\usepackage{url}
\usepackage{verbatim}
\usepackage{graphicx}
\usepackage{cite}
\usepackage{multirow}
\usepackage{booktabs}
\usepackage{color, xcolor}
\usepackage{enumitem}
\usepackage[
pdfauthor={derajan},
pdftitle={How to do this},
pdfstartview=XYZ,
bookmarks=true,
colorlinks=true,
linkcolor=blue,
urlcolor=blue,
citecolor=blue,
pdftex,
bookmarks=true,
linktocpage=true,   
hyperindex=true
]{hyperref}
\definecolor{yellow}{HTML}{E3A907}
\definecolor{green}{HTML}{4BA42C}
\definecolor{orange}{HTML}{C04F15}

\newcommand{\ar}[1]{{\color{black}#1}}
\newcommand{\arr}[1]{{\color{black}#1}}
\begin{document}

\title{Action Quality Assessment via Hierarchical Pose-guided Multi-stage Contrastive Regression}

\author{Mengshi Qi,~\IEEEmembership{Member,~IEEE,}
        Hao Ye,
        Jiaxuan Peng,
        Huadong Ma,~\IEEEmembership{Fellow,~IEEE} 
\thanks{This work is partly supported by the Funds for the NSFC Project (Grant 62202063/62572072/U24B20176), Beijing Natural Science Foundation (L243027). (\emph{Corresponding author: Mengshi Qi~(email:~qms@bupt.edu.cn)})}
\thanks{M. Qi, H. Ye, J. Peng and H. Ma are with State Key Laboratory of Networking and Switching Technology, Beijing University of Posts and Telecommunications, China.}
}
\markboth{IEEE Transactions on Image Processing}%
{Shell \MakeLowercase{\textit{et al.}}: A Sample Article Using IEEEtran.cls for IEEE Journals}


\maketitle

\begin{abstract}
\ar{Action Quality Assessment (AQA), which aims at the automatic and fair evaluation of athletic performance, has gained increasing attention in recent years.} However, athletes are often in rapid movement and the corresponding visual appearance variances are subtle, making it challenging to capture fine-grained pose differences and leading to poor estimation performance. Furthermore, most common AQA tasks, such as diving in sports, are usually divided into multiple sub-actions, each of which contains different durations. However, existing methods focus on segmenting the video into fixed frames, which disrupts the temporal continuity of sub-actions resulting in unavoidable prediction errors. To address these challenges, we propose a novel action quality assessment method through hierarchically pose-guided multi-stage contrastive regression. \ar{Firstly, we introduce a multi-scale dynamic visual-skeleton encoder to capture fine-grained spatio-temporal visual and skeletal features. Compared to mask or auxiliary visual features, skeletal features provide a more accurate representation during athletic movements. Then, a procedure segmentation network is introduced to separate different sub-actions and obtain segmented features.} Afterwards, the segmented visual and skeletal features are both fed into a multi-modal fusion module as physics structural priors, to guide the model in learning refined activity similarities and variances. Finally, a multi-stage contrastive learning regression approach is employed to learn discriminative representations and output prediction results. In addition, we introduce a newly-annotated FineDiving-Pose Dataset to improve the current low-quality human pose labels. In experiments, the results on FineDiving and MTL-AQA datasets demonstrate the effectiveness and superiority of our proposed approach. Our source code and dataset are available at https://github.com/Lumos0507/HP-MCoRe.
\end{abstract}


\section{Introduction}
\IEEEPARstart{A}{ction} quality assessment (AQA)~\cite{pan2019action,yu2021group,pirsiavash2014assessing,tang2020uncertainty,bai2022action,parmar2019and, shao2020intra} has garnered significant attention within the computer vision community, aiming to assess the quality of action execution. In practical applications, AQA plays a crucial role across various domains, such as automating evaluations in sports (\emph{e.g.,} gymnastics or diving)~\cite{zia2018video, parmar2019action, xu2019learning, yu2021group}, providing corrective guidance in rehabilitation training~\cite{zhang2015relative, sharma2014video, zia2015automated, zia2018video}, and obtaining performance feedback in skill learning. Unlike daily life videos, action videos in the AQA task are characterized by sequential processes and domain-specific execution standards, necessitating a deep understanding of actions and an accurate analysis of fine-grained sub-action features. These include subtle differences in limb and torso movements as well as variations in action durations. Additionally, performers in AQA tasks frequently exhibit rapid motion, which often leads to blurred frames, further complicating the assessment process.

Most existing approaches \cite{pan2019action,yu2021group,pirsiavash2014assessing,tang2020uncertainty,bai2022action} treat the AQA task as either a regression problem or a pairwise comparison regression problem, typically focusing on coarse-grained action features and utilizing holistic video representations extracted from videos. However, as previously discussed, coarse-grained features are not adequately supportive of high-quality evaluation, as holistic video representations struggle to capture the intricate variations between different sub-actions within a complete action sequence.

\ar{Although impressive performance has been achieved by using powerful visual backbones to extract video features in recent works \cite{du2023semantics,10737230,xu2024vision,majeedi2024rica2rubricinformedcalibratedassessment,xu2022finediving,qi2019stagnet,zhou2023hierarchical,qi2020stc,gedamu2023fine,an2024multi}, these methods still face significant challenges. For example, FineParser~\cite{xu2024fineparser} utilized human masks to guide the visual encoder in focusing on human-centric content; however, it lacks guidance at a finer granularity for capturing detailed action cues. Specifically, performers, particularly professional athletes in sports videos, often exhibit rapid motion, accompanied by complex limb coordination and bending, resulting in frequently blurred video frames. This poses challenges for backbones to accurately capture fine-grained pose and physical structure differences among various athletes performing the same action.}

\begin{figure}[t]
    \centering
    \includegraphics[width=0.5\textwidth]{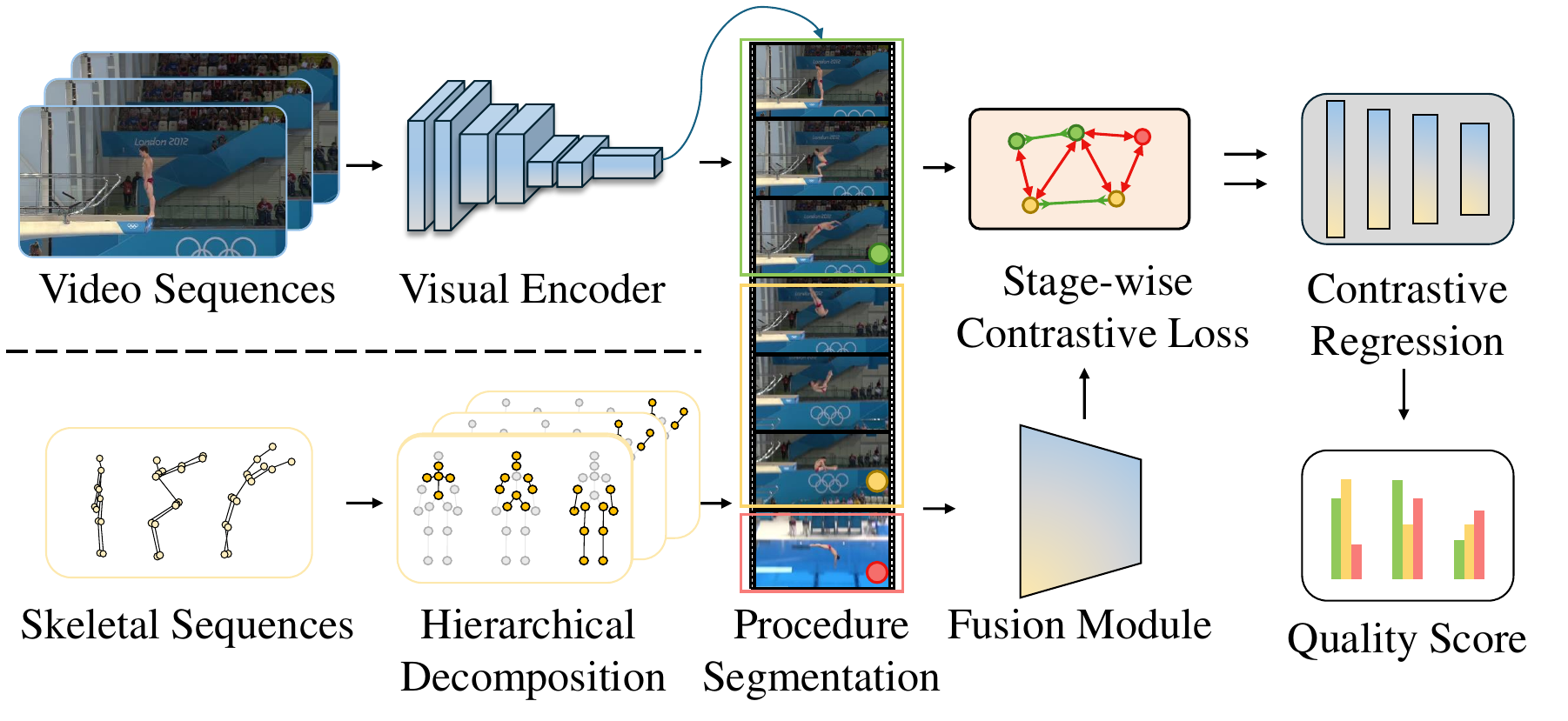} 
    \vspace{-7mm}
        \caption{Illustration of our proposed framework for AQA. The framework integrates both human visual and hierarchical skeletal information to capture fine-grained features and physical priors for the high-quality action assessment. Additionally, we introduce a procedure segmentation module that dynamically models the sub-action sequences. Subsequently, we fuse the skeletal and visual features to derive spatiotemporal features. Finally, we propose a contrastive learning-based regression approach to enhance the evaluation accuracy.
        } 
    \label{fig:framework}
    \vspace{-3mm}
\end{figure}

Additionally, sports such as diving and gymnastics often comprise multiple discrete phases within a single performance \cite{xu2022finediving}, exemplified by the \textit{take-off}, \textit{flight}, and \textit{entry} phases in a diving maneuver. Judges predominantly focus on the nuanced distinctions in athletes' executions across each sub-action, encompassing the number of flips, twists, and the posture during flight. Existing methods~\cite{xu2022finediving, zhang2023learning} typically segment input videos into fixed-frame clips to process these clips in a global manner. However, such approaches disrupt the temporal continuity of sub-actions, as the durations of phases often vary across different action categories. Identifying and segmenting each phase within an action sequence is believed to enable more precise analysis and conclusive evaluation \cite{pan2019action,yu2021group,li2022pairwise}. Therefore, it is crucial to dynamically segment the videos. However, clips corresponding to different sub-actions often share similar visual backgrounds and features, which presents significant challenges for dynamic action segmentation.

As illustrated in Figure \ref{fig:framework}, we propose a hierarchically pose-guided multi-stage action quality assessment framework to address the aforementioned challenges. This framework processes action videos and corresponding skeleton sequences as inputs. Compared to blurred video frames, the skeleton sequences of performers offer precise spatial coordinates and motion trajectories, thereby providing more fine-grained sub-action features \cite{nekoui2020falcons,qi2019attentive,lee2023hierarchically,lv2024disentangled,nekoui2021eagle}. The proposed framework primarily consists of two branches: a static branch and a dynamic branch. The static branch utilizes a static visual encoder to process video frames, aiming to retain more contextual details and enhance action representation. The proposed dynamic branch, comprising a dynamic visual encoder and a hierarchical skeletal encoder, captures fine-grained spatiotemporal differences and physical priors to guide feature fusion through a multi-modal fusion module. During this process, the procedure segmentation module distinguishes different stages of sub-actions and segments all features accordingly. \ar{Finally, the multi-stage features from both the static and dynamic branches are passed through the stage contrastive regression module and the splash regression module, to extract discriminative features and emphasize splash-related cues, leading to an understanding of differences in athlete actions across different videos. To optimize the model's understanding of different sub-actions, we design a stage contrastive loss to perform unsupervised training and learn the differences between sub-actions.}

\ar{It should be noted that this paper is an extension of our conference paper \cite{an2024multi}. Compared to the preliminary version, we leverage an additional skeletal modality to obtain hierarchical human pose features. Given the limitations in existing datasets characterized by the poor quality or absence of skeletal labels, we also present a newly-annotated FineDiving-Pose Dataset with refined pose labels, which are collected through a combination of manual annotation and automatic generation to boost the related field further. Furthermore, we propose a multi-modal fusion module to integrate visual features and skeletal features and add a static branch to capture human static features. Additionally, we note that the splash appearance is a critical cue in diving score assessment. Thus, we incorporate a splash detection module and introduce a corresponding splash loss during training. This loss guides the model to pay greater attention to splash-related visual patterns during water entry, thereby improving the accuracy of score prediction. Finally, we perform additional experiments on another public benchmark dataset (MTL-AQA\cite{parmar2019and}), and we conduct more detailed ablation experiments and then present more qualitative results to demonstrate the effectiveness of each component within our proposed framework.}
 
The contributions of this paper are summarized as follows:
\begin{itemize}
\item We propose a hierarchical pose-guided AQA framework, which introduces the combination of static and dynamic visual branches to decompose pose information and guide the fusion of pose features with dynamic video features for capturing fine-grained spatiotemporal features.
\item We develop a procedure segmentation network, which divides the input sequences into multiple stages to learn broader contextual information, along with a stage contrastive learning regression module to enhance the model's ability to learn differences between sub-actions.
\item We introduce a newly annotated \emph{FineDiving-Pose Dataset} to boost the AQA research further, which contains 12,722 human annotated pose labels and 288,000 automatically-annotated pose labels by our proposed annotated pipeline. 
\item We conduct extensive experiments on two mainstream datasets, \emph{FineDiving} \cite{xu2022finediving} and \emph{MTL-AQA} \cite{parmar2019and}, to evaluate our method, demonstrating that our approach outperforms state-of-the-art methods.
\end{itemize}

\begin{figure*}[t]
  \centering
  \includegraphics[width=0.9\linewidth]{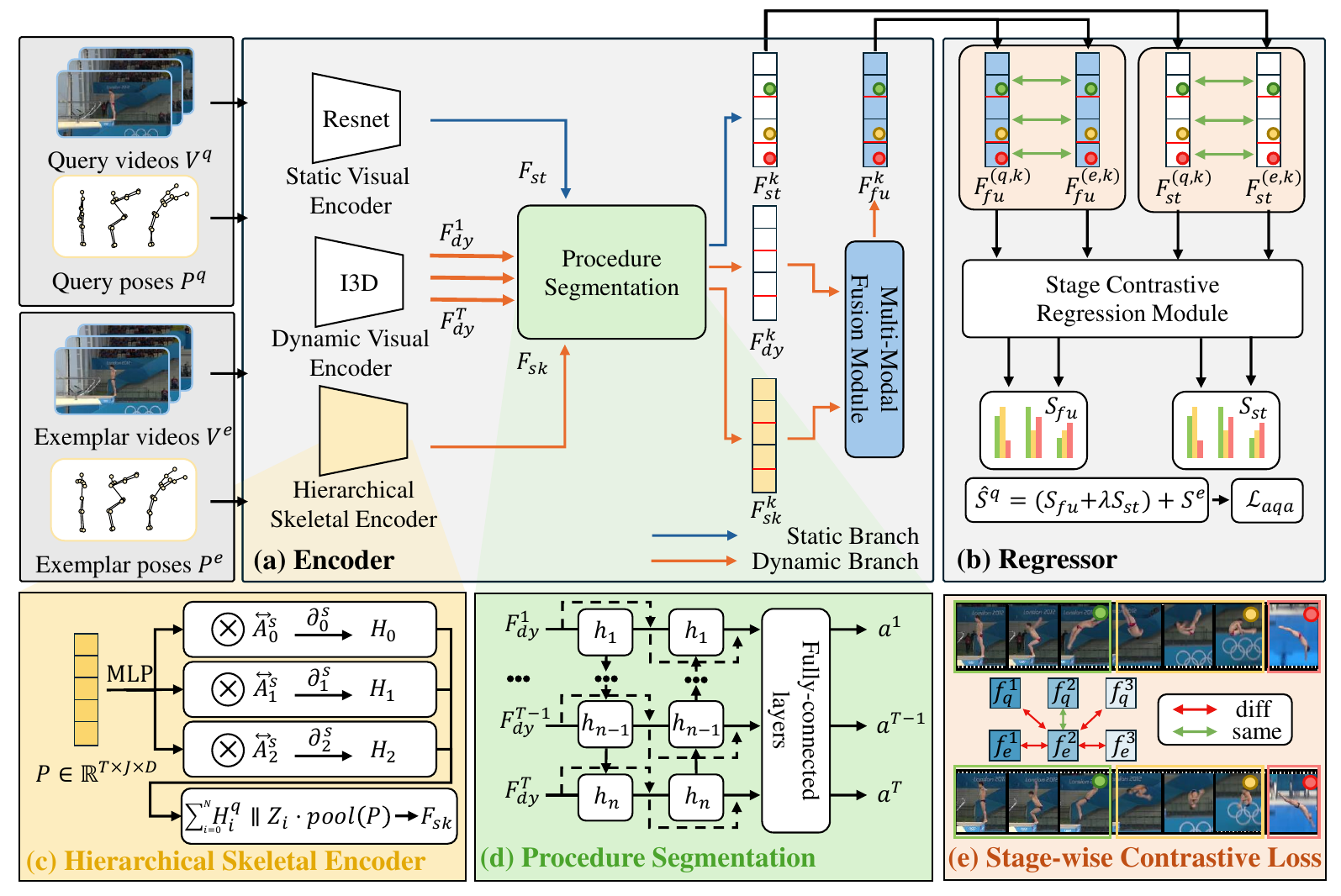}
  \vspace{-2mm}
  \caption{ Overview of our proposed hierarchical pose-guided multi-stage action quality assessment framework. \textbf{Encoder:} The framework takes query pairs ${V^q, P^q}$ as input for testing, while exemplar pairs ${V^e, P^e}$ are selected from an existing dataset. The dynamic visual encoder and the \textcolor{yellow}{(c) hierarchical skeletal encoder} capture the spatiotemporal visual and pose features $F_{dy}$ and $F_{sk}$. The static visual encoder captures static human features $F_{st}$. The \textcolor{green}{(d) procedure segmentation network} segments these features into $\mathbb K$ stages, resulting in $F_{dy}^k$, $F_{sk}^k$ and $F_{st}^k$. The \textcolor{orange}{(e) stage-wise contrastive loss} is applied to enhance segmentation accuracy. The multi-stage features of $F_{dy}^k$ and $F_{sk}^k$ are input into the multi-modal fusion module to obtain the fused features $F_{fu}^k$. \textbf{Regressor:} \ar{The inputs are fused dynamic features and static features of the $k$-th segmentation, denoted as $F_{fu}^{(q,k)}$, $F_{st}^{(q,k)}$, $F_{fu}^{(e,k)}$ and $F_{st}^{(e,k)}$.} The static features and fused features are fed into the stage contrastive regression module separately to obtain the relative scores $S_{fu}$ and $S_{st}$. Finally, a confidence value $\lambda$ is applied, and the exemplar score $S_e$ is added to produce the predicted score $\hat{\mathit{S^q}}$.}
  \label{fig:NetWorkArchitecture}
\end{figure*}

\section{RELATED WORK}
\noindent\textbf{AQA.}~\ar{Recent comprehensive surveys~\cite{zhou2024comprehensive,yin2025decade} systematically evaluated the progress and future directions of Action Quality Assessment. Zhou~\emph{et al.}~\cite{zhou2024comprehensive} reviewed over 150 publications and proposed a modality-based hierarchical taxonomy, along with a unified benchmark to facilitate fair comparison across methods. In parallel, Yin~\emph{et al.}~\cite{yin2025decade} conducted a systematic review of more than 200 papers, identifying key challenges and emerging trends in the AQA field. }~Currently, the AQA task mainly follows two types of formulations: \emph{regression} and \emph{pairwise ranking}.~\emph{Regression-based} approaches~\cite{pirsiavash2014assessing,parmar2019and,zeng2020hybrid,tang2020uncertainty,xu2022finediving,zhou2023hierarchical} are widely applied in sports such as diving, skiing, and synchronized swimming. \arr{Pirsiavash~\emph{et al.}~\cite{pirsiavash2014assessing} pioneered the use of regression models for human action quality assessment, predicting action scores directly from video features.} 
Parmar~\emph{et al.}~\cite{parmar2019and} proposed a multi-task clip-level scoring method by utilizing spatiotemporal action features. Tang~\emph{et al.}~\cite{tang2020uncertainty} proposed an uncertainty-aware score distribution learning framework, thereby addressing the uncertainties arising from subjective evaluations. Zeng~\emph{et al.}~\cite{zeng2020hybrid} proposed a hybrid method that integrates static and dynamic action features, accounting for the contributions of different temporal segments. Xu~\emph{et al.}~\cite{xu2022finediving} proposed a procedure-aware representation by designing a temporal segmentation attention module. Zhou~\emph{et al.}~\cite{zhou2023hierarchical} proposed a hierarchical graph convolutional network (GCN) to refine semantic features, and aggregate dependencies for analyzing action procedures and motion units. While~\emph{Pairwise ranking-based} approaches~\cite{li2019manipulation, yun2025semi,qi2021semantics,doughty2019pros,pan2021adaptive} tackle the challenge of distinguishing subtle differences between actions in similar contexts, by evaluating the score of a given video relative to other videos through pairwise comparisons. Yu~\emph{et al.}~\cite{yu2021group} were the first to propose a pairwise ranking-based approach for diving to learn relative scores. Li~\emph{et al.}~\cite{li2022pairwise} further enhanced the model's ability to capture subtle score differences by incorporating Pairwise Contrastive Learning Network. Similarly, An~\emph{et al.}~\cite{an2024multi} proposed a multi-stage contrastive regression framework to efficiently extract spatiotemporal features for AQA. \ar{
\arr{Xu~\emph{et al.}~\cite{xu2024procedure,10946879} proposed a spatial-temporal segmentation attention to parse actions and employ procedure-aware cross-attention to improve scoring accuracy.} An~\emph{et al.}~\cite{an2024multi} proposed a multi-stage contrastive regression framework to efficiently extract spatiotemporal features for AQA. Gedamu~\emph{et al.}~\cite{10706814} proposed a self-supervised sub-action parsing network with a teacher-student architecture to enable semi-supervised AQA.} These methods primarily focus on evaluating entire video sequences either within the same video or across different videos. \arr{In contrast, our approach divides video sequences into different phases and then uses the fused visual-skeletal representation, enabling fine-grained contrastive learning of motion differences at each stage. }

\noindent\textbf{Multimodal Learning in AQA.}~\ar{In a recent survey paper, Liu~\emph{et al.}~\cite{liu2024vision} emphasized that the current AQA methodologies can be broadly classified into two mainstream paradigms: skeleton-based and vision-based approaches. \arr{\emph{Vision-based} approaches typically leverage powerful backbone architectures, such as C3D~\cite{Tran2015Learning} and I3D~\cite{carreira2017quo}. Parmar~\emph{et al.}~\cite{parmar2017learning} were the first to utilize a segment-level C3D to extract features, with an LSTM network~\cite{shi2015convolutional} to capture temporal relationships between video segments.} Building on this, Li~\emph{et al.}~\cite{li2019manipulation} introduced a spatial attention network, which involved segmenting the video, performing random sparse sampling, and using RGB images and optical flow to extract informative features. Wang~\emph{et al.}~\cite{wang2021tsa} subsequently integrated a single-object tracker to enhance the distinction between foreground and background in feature maps.}~\emph{Skeleton-based}~approaches~\cite{ahmadi2009towards,bourgain2018effect,pan2019action,nekoui2021eagle,lee2023hierarchically,okamoto2024hierarchical} focused on leveraging detailed human pose information to capture the positions and movements of individual body parts. Early approaches~\cite{ahmadi2009towards,bourgain2018effect} relied on specialized equipment to capture athlete poses, which limited their applicability to real-world scenarios. However, recent advances in pose estimation have rendered these approaches more practical. Pan~\emph{et al.}~\cite{pan2019action} proposed to learn interactions between pose joints, demonstrating that fine-grained pose information can enhance model performance. Nekoui~\emph{et al.}~\cite{nekoui2021eagle} employed a joint coordination evaluator in combination with multi-scale convolution to capture temporal dependencies, while okamoto~\emph{et al.}~\cite{okamoto2024hierarchical} conducted a detailed analysis of athletes' actions using a neuro-symbolic approach. In our work, we propose a new hierarchical pose-guided action quality assessment method, which decomposes athlete skeletal poses into several meaningful sub-sets to highlight diverse motion features, thereby enhancing AQA performance.

\section{PROPOSED APPROACH}
\label{sec:format}
\subsection{Overview}
Our proposed framework is illustrated in Figure \ref{fig:NetWorkArchitecture}. Firstly, a multi-scale visual-skeletal encoder is introduced to capture fine-grained spatiotemporal and static features from visual frames and skeletal pose. Next, the procedure segmentation module is designed to distinguish different temporal stages for both dynamic and static features. By comparing the action features across different stages, stage-level features are obtained. The segmented dynamic visual features and skeletal features are then processed through the multi-modal fusion module. Finally, a multi-stage contrastive learning regression method is utilized to learn discriminative representations to estimate the quality score. 

\textbf{\emph{Problem Formulation}.} Given a pair of query video $V^{q}$ and exemplar video $V^{e}$, along with their respective skeletal sequences $P^{q}$ and $P^{e}$, the object is to predict quality score $\hat{S}^{q}$ of $V^{q}$ based on the quality score $S^{e}$ of $V^{e}$, which is formulated as the following:
\begin{equation}
    \hat{S}^{q}=\mathcal{F}(V^{q},P^{q},V^{e},P^{e}|\Theta)+S^{e},
\end{equation}
where $\mathcal{F}(\cdot|\Theta)$ represents the overall network architecture, with $\Theta$ denoting the parameters of the model, $V\in\mathbb{R}^{T\times H\times W\times C}$ and $P\in\mathbb{R}^{T\times J\times D_c}$, where $T,H,W,C$ denote the number of frames, height, width, and the number of channels respectively, and $J,D_c$ refer to the number of human joints and the coordinate dimension of joints. This regression problem aims to obtain the final score by predicting the relative score difference between the query video and the exemplar video.

\begin{figure}[t]
    \centering
    \includegraphics[width=0.45\textwidth]{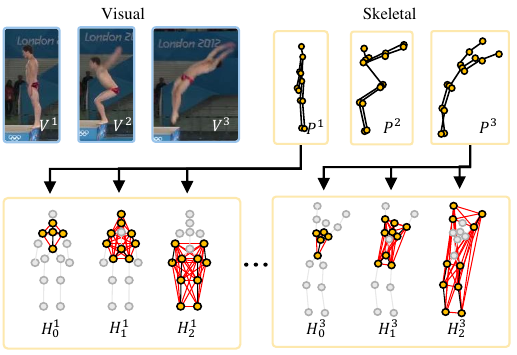} 
    \vspace{-2mm}
    \caption{The illustration of the hierarchical pose encoding. We decompose the human pose into three levels using a graph convolutional network. The first level focuses on the torso, capturing the correctness of the torso rotation. The second level focuses on the inner parts of the limbs, capturing transitions between actions. The third level targets the outer parts of the limbs, capturing the coordination of the athlete's body movements.}
    \label{Hierarchical Decomposed Pose}
    \vspace{-3mm}
\end{figure}

\subsection{Multi-Scale Visual-Skeletal Encoder}
As illustrated in Figure \ref{fig:NetWorkArchitecture}, we introduce the multi-scale visual-skeletal encoder that consists of three components: static visual encoder, dynamic visual encoder for extracting visual features, and the hierarchical skeletal encoder for capturing human pose features.

\textbf{\emph{Static Visual Encoder}:} This component is designed to capture human static features and enhance the action representation at each temporal step. We adopt ResNetY \cite{he2016deep} as the backbone, which places more emphasis on appearance information from individual RGB frames, compared to I3D \cite{carreira2017quo}. The encoder consists of a ResNetY model denoted as ${\mathcal R}$ and a multi-branch downsampling module $Down_{mul}$. Specifically, the ResNet model is primarily adopted to extract high-level global features from the input video, while the multi-branch downsampling module, consisting of a convolutional layer and three downsampling layers, captures local detailed information at different spatial scales. The static feature $F_{st}$ can be obtained from this module as follows:
\begin{equation}
    F_{st}=Down_{mul}\big(\mathcal R(V)\big), 
\end{equation}
where $V$ represents query or exemplar video frames.

\textbf{\emph{Dynamic Visual Encoder}:} This component is responsible for extracting spatiotemporal features and complex dynamic information from the video sequence, such as the sequence of an action from preparation to take-off. We adopt I3D \cite{carreira2017quo} as the backbone network, denoted as $\mathcal{I}$, to encode the video input. To expand the temporal dimension's receptive field, mixed convolution layers are introduced before the max-pooling layer. 
Specifically, the query video $V^q$ is fed into the backbone network $\mathcal I$, followed by convolution operations along the temporal axis to dilate the temporal dimension of $\mathcal I(V^{q})$, and a max-pooling operation along the spatial axis, which is formulated as follows:
\begin{equation}
    F_{dy}^{q}=maxpool(Conv_{mix}(\mathcal I(V^{q}))),
\end{equation}
where $Conv_{mix}$ represents the mixed convolution layers, consisting of three layers, and $F_{dy}^{q}\in\mathbb{R}^{T\times D}$ is the final visual spatiotemporal feature. Similarly, the exemplar video $V^e$ also obtains the corresponding $F_{dy}^e$.

\textbf{\emph{Hierarchical Skeletal Encoder}:} Taking into account the kinematic principles and trajectory patterns, we observe that distant joints (e.g., hands and feet) execute both substantially and refined actions, while central joints (e.g., the torso) exhibit minimal actions yet are highly responsive to the overall body displacement. In light of these observations, we endeavor to decompose the full range of skeletal joints into three distinct semantic levels, differentiated by the amplitude and directionality of joint movements, as depicted in Figure~\ref{Hierarchical Decomposed Pose}.

To be specific, the lowest level $H_0$ encapsulates the motion of the torso, which is indicative of the body's spatial dynamics within pose sequences. The highest level $H_2$, characterized by the highest motion amplitude, captures the intricate and rich movements, encompassing joints that engage in agile and rapid sub-actions, such as the swift movements of the wrists and ankles observed in twists in straight, pike or tuck positions within diving scenario. The features derived from $H_2$ are particularly fine-grained, as these motions are the focus of scrutiny during assessment. The intermediate level $H_1$ serves as a bridge between $H_0$ and $H_2$, tasked with capturing the transitional aspects of sub-actions, exemplified by the changes in the angles of the elbow and knee from the \textit{take-off} to the \textit{flight} in diving scenario.

As illustrated in Figure \ref{fig:NetWorkArchitecture} (c), the hierarchical skeletal encoder enhances feature capture capabilities at different levels through four parallel branches, including three graph convolutions. Each graph convolution corresponds to a specific skeletal level (e.g., $H_0$). For each level, we perform GCN operations to extract feature and concatenate the features of each level, which is formulated as
\begin{equation}
    H^q_i=\parallel_{s\in S}\{\overleftrightarrow{\textbf{A}_i^s}\,\text{MLP}(P^q)\,\partial_i^s\},
\end{equation}
where $\overleftrightarrow{\textbf{A}_{i}^{s}}\in\mathbb{R}^{N_S \times D\times D}$ denotes the skeletal node adjacency matrix for the $i$-th layer, and $S = \{s_{id},s_{cf},s_{cp}\}$ denotes above-mentioned three pose skeleton subsets, and $s_{id}$,$s_{cf}$,$s_{cp}$ indicate identity, centrifugal, and centripetal joint subsets, respectively.  $\text{MLP}(\cdot)$ represents one layer MLP with parameters $W\in\mathbb{R}^{D^\prime\times D}$, $\partial_{i}^{s}$ is the point-wise convolution operation, $\parallel$ denotes concatenation along with the channel dimension, and $H^q=\{H_{0}^q, H_{1}^q, H_{2}^q\}$.

The three semantic levels defined above are introduced to model kinematic and motion information, but they exist as sparse graphs where the edges represent only physical connections. These edges are inadequate to capture implicit features that are embedded within the relationships among distant joint nodes. Therefore, we expand the human pose graph through connecting all nodes in adjacent levels. Specifically, we first apply average pooling to obtain the average feature of each node. Then, the EdgeConv \cite{dgcnn} is employed to capture both explicit and implicit features. This operation effectively reflects relationships between physically joint edges in shallow layers and semantically similar edges in deeper layers. The EdgeConv operation can be defined as follows:
\begin{equation}
    F_{sk}^q=\sum_{i=1}^N[H^q_i \parallel Z_i\cdot \frac{1}{T}\sum_{t=1}^T\text{MLP}(P^q)],
\end{equation}
where $Z_i$ is the EdgeConv operation at the $i$-th layer, $N$ denotes different skeletal semantic levels. The skeletal pose $P^q$, after passing through the hierarchical skeletal encoder, produces the skeletal feature output $F_{sk}^{q}\in\mathbb{R}^{T\times J\times D}$. Similarly, $P^e$ can also obtain the skeletal feature output $F_{sk}^e$.

\begin{figure}[t]
    \centering
    \includegraphics[width=0.45\textwidth]{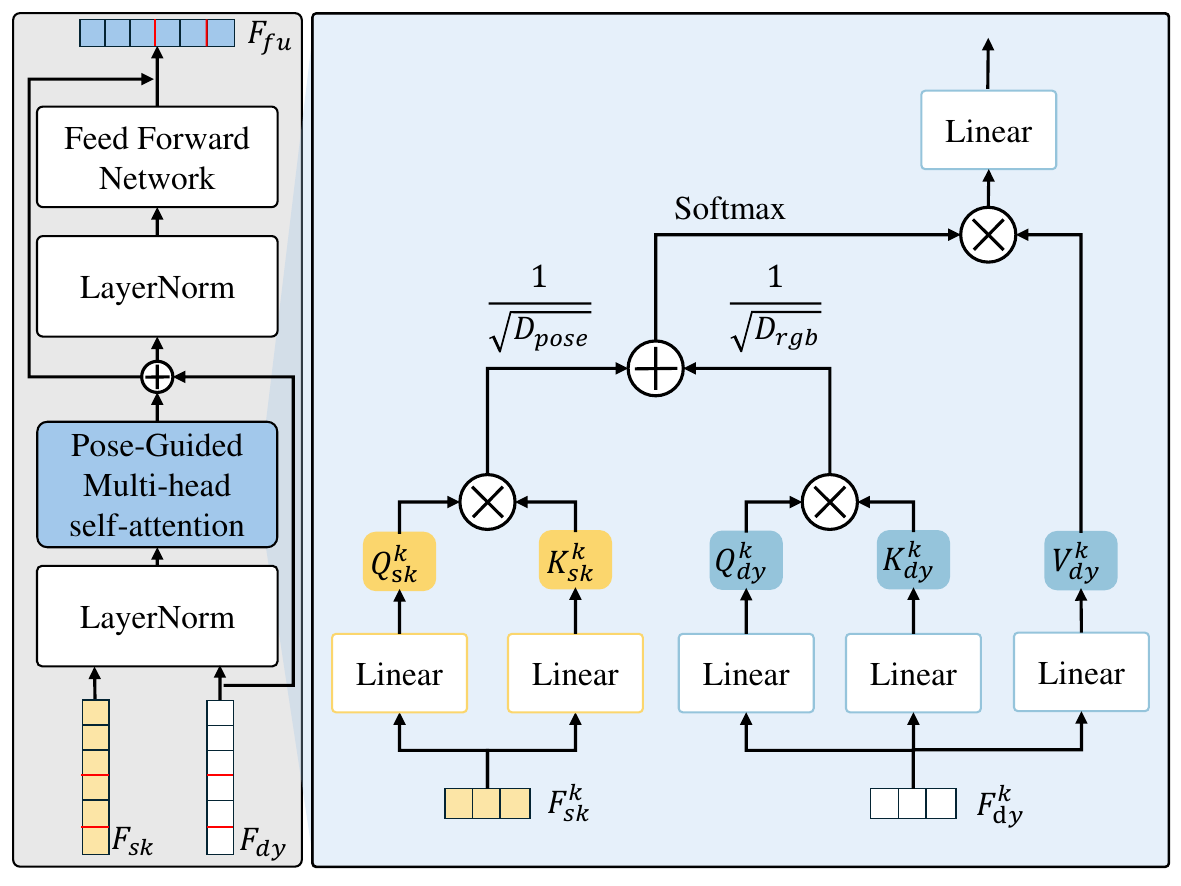}
    \vspace{-3mm}
    \caption{Illustration of the Multi-modal Fusion Module. We propose a pose-guided attention into such module where the input pose and visual features share the same procedure segmentation information, ensuring consistency in their corresponding spatiotemporal information.}
    \label{Fusion Module}
    \vspace{-4mm}
\end{figure}

\subsection{Procedure Segmentation Module}
During competition, judges typically focus on the performance of the athlete at different stages of the action \cite{okamoto2024hierarchical}, assigning an overall score based on the effectiveness of each sub-action. For example, the athlete’s movement can be divided into \textit{take-off}, \textit{flight}, and \textit{entry} in the diving scenario. In our approach, we independently score the actions within each stage to achieve more accurate predictions. To enable the model to distinguish the significant variances between sub-actions, we propose a procedure segmentation network. We assume that the action video can be divided into $\mathbb K$ stages, and the goal of the procedure segmentation network is to predict the $\mathbb K-1$ moments where stage transitions occur. To capture long-term temporal dependencies, we propose a procedure segmentation network $\mathcal P$ based on the Bidirectional Gated Recurrent Unit (Bi-GRU)~\cite{cho2014learning} followed by a fully connected layer and a softmax operation. Specifically, the dynamic visual feature $F_{dy}$ is input to $\mathcal P$ to predict the stage transition labels, which can be expressed as  follows:
\begin{equation}\mathcal P(F_{dy})=[\boldsymbol{\hat{a}_{1}},\boldsymbol{\hat{a}_{2}},\ldots \boldsymbol{\hat{a}_{T}}].
\end{equation}
Here, $\boldsymbol{\hat{a}_{i}}$ denotes the predicted label of the $i$-th frame. If the predicted labels of the $i$-th frame and the $(i-1)$-th frame are identical, it indicates that both frames belong to the same sub-action stage, and we set $\boldsymbol{\hat{a}_{i}} = 0$. Conversely, if the predicted labels differ, it signifies that the $i$-th frame marks the transition to the next sub-action stage, and set $\boldsymbol{\hat{a}_{i}} = 1$.

Next, we identify the frames with the highest probability as the prediction of stage transitions. The per-frame classification task is optimized using the cross-entropy loss. 
The loss function $\mathcal{L}_{ce}$ calculates the difference between the predicted stage transition moments and the ground truth, which can be formulated as follows:
\begin{equation}
\mathcal{L}_{ce} = \sum_{i=1}^\mathbb K \text{CE}\left(\boldsymbol{a_i},\boldsymbol{\hat{a}_i}\right),
\end{equation}
where $a_i$ denotes the ground truth of the stage transition label.

\subsection{Multi-Modal Fusion Module}
To integrate the visual spatiotemporal feature $F_{dy}$ and the hierarchical skeletal feature $F_{sk}$ together, we propose a novel multi-modal fusion module, as illustrated in Figure \ref{Fusion Module}. The attention map from $F_{sk}$ is added to the map of $F_{dy}$ to enhance the model's ability to capture and understand human poses, ensuring accurate comprehension of target features across different perspectives or pose variations. 

Specifically, the details of the fusion module are formulated in Equations \ref{equation_transfomer1} and \ref{equation_transfomer2}. First, the $i$-th stage skeletal features $F_{sk}^i$ and visual features $F_{dy}^i$ within the $\mathbb K$ stages are fed into individual linear layers to yield the queries $Q^i$, keys $K^i$, and values $V^i$ with the same dimensions. The attention maps for the visual and skeletal features are then calculated separately and element-wise added to form a pose-guided attention map $\alpha^{i}$. The pose-guided attention map is then multiplied by the value $V^i$ to obtain the attention output $Atten^i$:
\begin{equation}
    \alpha^{i}=\mathrm{Softmax}\left(\frac{K^{i}_{dy}\cdot Q^{i}_{dy}}{\sqrt{D_{dy}}}+\rho \frac{K^{i}_{sk}\cdot Q^{i}_{sk}}{\sqrt{D_{pose}}}\right),
\label{equation_transfomer1}
\end{equation}
\begin{equation}
    Atten^i=Multihead(V^i_{dy}\cdot\alpha^i).
\label{equation_transfomer2}
\end{equation}
where $\rho$ represents a learnable weight parameter, $D_{dy}$ and $D_{pose}$ are the dimension values of the corresponding features, $i\in[1,\mathbb K]$. The whole formulation of the multi-modal fusion can be expressed as follows:
\begin{equation}
    H_{dy} = Atten^i+LN(F^i_{dy}),
\end{equation}
\begin{equation}
    \mathrm{F}_{\mathrm{fu}}^i=FFN(LN(H_{dy}))+H_{dy},
\end{equation}
where $LN$ and $FFN$ denote Layer Normalization and Feed Forward Network, respectively.

\subsection{Multi-stage Contrastive Regression Module}
\textbf{\emph{Stage-wise Contrastive Loss}:} After obtaining the segmented fused features $F_{fu}$ and static features $F_{st}$, we introduce a stage-wise contrastive loss, denoted as \(\mathcal{L}_{\text{cont}}\), to further enhance the differentiation between sub-actions at various stages. We apply the stage-wise contrastive loss to $F_{fu}$ and $F_{st}$ separately, denoted as $F$ in the following for simplification.

Formally, we define a critic function \(sim\left(F^{(q,k)},F^{(e,k)}\right) \), where \({cos}\) denotes cosine similarity, and \({norm}\) denotes a normalization function designed to enhance the critic's expressive capacity, as follows:
\begin{equation}
\boldsymbol{\delta}\left(F^{(q,k)},F^{(e,k)}\right) = \cos{(norm(F^{(q,k)}),norm(F^{(e,k)}))}.
\end{equation}

During the calculation of the contrastive loss,  we treat the features of the same stage in both the query video and the sample video as positive pairs, denoted as $\boldsymbol{\epsilon}(F^{(q,k)},F^{(e,k)})$. 
While negative pairs $\boldsymbol{\zeta}(F^{(q,k)},F^{(e,k)})$ can be divided into inter-video or intra-video pairs. Inter-video negative pair $(F^{(q,k)},F^{(e,l)})$ consists of features from different stages in different videos, while intra-video pair $(F^{(q,k)},F^{(q,l)})$ refers to features from different stages within the same video. 
The positive pairs $\boldsymbol{\epsilon}(F^{(q,k)},F^{(e,k)})$ and negative pairs $\boldsymbol{\zeta}(F^{(q,k)},F^{(e,k)})$ can be formulated as follows:

\begin{equation}
\boldsymbol{\epsilon}(F^{(q,k)},F^{(e,k)}) = e^{\boldsymbol{\delta}(F^{(q,k)},F^{(e,k)}) / \tau},
\end{equation}
\begin{equation}
    \boldsymbol{\zeta}(F^{(q,k)},F^{(e,k)})=\sum_{k=1;k\neq l}^\mathbb K(\underbrace{e^{\frac{\boldsymbol{\delta}(F^{(q,k)},F^{(e,l)})}{\tau}}}_{\text{inter negative pairs}}+\underbrace{e^{\frac{\boldsymbol{\delta}(F^{(q,k)},F^{(q,l)})}{\tau}}}_{\text{intra negative pairs}}),
\end{equation}
where $\tau$ denotes the temperature parameter and we set as 0.5 in practice.
We define the pairwise objective for pair  $\left(F^{(q,k)},F^{(e,k)}\right)$ with the positive and negative terms as:
\begin{equation}
\ell (F^{(q,k)},F^{(e,k)})= 
\log \frac{
\boldsymbol{\epsilon}(F^{(q,k)},F^{(e,k)})
}
{
\boldsymbol{\epsilon}(F^{(q,k)},F^{(e,k)})
+
\boldsymbol{\zeta}(F^{q,k)},F^{(q,k)})
}.
\end{equation}

The stage contrastive loss is defined as the average over all given video pairs, denoted as $\mathcal{L}_{cont}$, formulated as follows:
\begin{equation}
\mathcal{L}_{cont}=\frac{1}{2 \mathbb K} \sum_{i=1}^\mathbb K\left[\ell (F^{(q,k)},F^{(e,k)})+\ell (F^{(e,k)},F^{(q,k)})\right].
\end{equation}

\textbf{\emph{Score Regression}:} We leverage the powerful representation capability of the Transformer to capture differences between the query pairs and the exemplar pairs across different stages. Specifically, we utilize a Transformer decoder $\mathcal D$ to calculate the differences between the query and exemplar features for the $k$-th stage:
\begin{equation}
    f_{fu}^{k}=\mathcal D\left(F_{fu}^{(q,k)},F_{fu}^{(e,k)}\right),
\end{equation}
\begin{equation}
    f_{st}^k=\mathcal D\left(F_{st}^{(q,k)},F_{st}^{(e,k)}\right),
\end{equation}
where $\mathcal D$ in both branches shares the same structure but different parameters, $f_{fu}^{k}$ and $f_{st}^k$ represent fusion and static difference features, respectively. In each block of $\mathcal D$, cross-attention is adopted to calculate the differences between query $F^{(q,k)}$ and exemplar $F^{(e,k)}$.  
$F_{fu}^{(q,k)}$ refers to the fusion feature of the $k$-th stage of the query video. 

Finally, based on the generated difference fusion and static features, our stage contrastive regression network quantifies the action score difference between query and exemplar videos by learning relative scores. The predicted score $\hat{S}^q$ for the query video $V^{q}$ is calculated based on the exemplar video $V^{e}$ as follows:
\begin{equation}
    \hat{S}^q=S^e+\mathcal{CR}(F_{fu}, F_{st}),
\end{equation}
\begin{equation}
    \mathcal{CR}(F_{fu}, F_{st}) =\sum_{k=1}^\mathrm{\mathbb K}\lambda_k(\mathrm{\mathcal M}_{\mathrm{fu}}(f_{fu}^k)+\mathrm{\mathcal M}_{\mathrm{st}}(f_{st}^k)),
\end{equation}
where $F_{fu}=\big[f_{fu}^1, \cdots, f_{fu}^\mathbb K\big]$ and $F_{st}=\big[f_{st}^1, \cdots, f_{st}^\mathbb K\big]$, $\mathcal{CR}$ denotes the stage-wise contrastive regression, which calculates the relative score difference between query video and exemplar video. This module incorporates two multi-layer perceptrons (MLPs) with ReLU activation, denoted as $\mathcal M_{fu}$ and $\mathcal M_{st}$ respectively. $\lambda_k$ is a learnable weight parameter for different stages, which is normalized to ensure the contributions of the various stages are appropriately balanced.  $S^e$ is the actual score of the exemplar video, and $\mathbb K$ is the number of stages. 

~\ar{In addition, we observed that existing methods often fail to adequately model splash-related cues during the water entry phase. To enhance the model's sensitivity to this critical visual signal, we introduce a splash detection~\cite{wu2019detectron2} that identifies splash regions in each video frame and records their size, shape, and spatial location. We then integrate the splash area over time to compute a video-level splash score as the ground-truth, defined as follows:}
\ar{
\begin{equation}
\texttt{$S_{splash}$} = \int \mathbf{s}(t) \, dt \approx \sum_{i=1}^{N-1} \frac{\mathbf{s}_i + \mathbf{s}_{i+1}}{2} \cdot \Delta t ,
\end{equation}
}
\noindent \ar{where $s(t)$ denotes the splash area as a function of time, $\mathbf{s}_i$ denotes the splash area in frame $i$, $\Delta t$ is the time step, and $N$ is the number of frames in which a splash is present.}

\ar{To incorporate this cue into the learning process, we feed both the fused and static features into a dedicated splash regression module $\mathcal{SR}$. This module also incorporates two multi-layer perceptrons (MLPs) with ReLU activation, denoted as $\mathcal A_{fu}$ and $\mathcal A_{st}$ respectively. The module is trained to minimize the discrepancy between the predicted splash score and the computed ground-truth splash score, as follows:}
\ar{
\begin{equation}
    \hat{S}^q_{splash}=S^e_{splash}+\mathcal{SR}(F_{fu}, F_{st}),
\end{equation}
\begin{equation}
    \mathcal{SR}(F_{fu}, F_{st}) =\sum_{k=1}^\mathrm{\mathbb K}\lambda_k(\mathrm{\mathcal A}_{\mathrm{fu}}(f_{fu}^k)+\mathrm{\mathcal A}_{\mathrm{st}}(f_{st}^k)),
\end{equation}
}
\ar{\noindent where $\hat{S}^q_{splash}$ denotes the predicted splash score, $S^e_{splash}$ is the ground-truth splash score of video $V^e$.}

\ar{To evaluate the accuracy of score prediction in the AQA task, we utilize the Mean Squared Error (MSE) as a metric. The MSE calculates the squared difference between the predicted scores and the ground truth values as the following:}
\ar{\begin{equation}
    \mathcal{L}_{aqa}=\mathrm{MSE}(S^{q},\hat{S}^{q}),
\end{equation}
\begin{equation}
    \mathcal{L}_{splash}=\mathrm{MSE}(S_{splash}^{q},\hat{S}_{splash}^{q}),
\end{equation}
\begin{equation}
    \mathrm{MSE}=\frac{1}{N}\sum_{i=1}^{N} (y_{i}-\hat{y}_{i})^{2},
\end{equation}
where $N$ is the number of samples, $y_i$ is the true value of the $i$-th sample, and $\hat{y}_i$ is the predicted value. }

\subsection{Optimization and Inference}
\textbf{\emph{Optimization}.} 
\ar{For each video pair in the training data with pose sequences and score label $S^q$, the overall objective function for our task can be written as:
\begin{equation}
\mathcal{L}=\mathcal{L}_{aqa}+\mathcal{L}_{splash}+\mathcal{L}_{ce}+\mathcal{L}_{cont}.
\end{equation} }
\textbf{\emph{Inference}.}
For the query video $V^{q}$ and the corresponding skeletal pose $P^{q}$ in the test set, we adopt a multi-sample voting strategy to select $N$ samples from the training set, obtaining the corresponding exemplar videos $\{V_i^e\}_{i=1}^N$ and exemplar skeletal pose $\{P_i^e\}_{i=1}^N$. We take  $\{(V^q,P^q,V_i^e,P_i^e)\}_{i=1}^N$ as inputs, with their corresponding scores $\{S_i^e\}_{i=1}^N$. The inference process can be formulated as follows:
\begin{equation}
    \hat{S}^q=\frac{1}{N}\sum_{i=1}^N(\mathcal{F}(V^q,P^q,V_i^e,P_i^e|\Theta)+S_i^e).
\end{equation}

\section{FINEDIVING-POSE DATASET}
To enable studies on fine-grained articulated action, we construct a new dataset termed FINEDIVING-POSE, which includes fully-annotated human skeletal pose labels.

\subsection{Data Source}

The widely-adopted FineDiving Dataset \cite{xu2022finediving} and MTL-AQA Dataset \cite{parmar2019and} both lack human pose labels. The FineDiving Dataset contains 52 different diving categories, which substantially overlap with the categories present in the MTL-AQA Dataset. Therefore, we source data from the FineDiving dataset \cite{xu2022finediving} and annotate all samples in FineDiving. A total of 367 videos and 12,722 frames are manually labeled to extract paired 2D human pose keypoints and bounding box labels according to the raw resolution, without any cropping or scaling. Additionally, using an automated annotation method, we label 3,000 videos and 288,000 frames, each accompanied by annotated 2D keypoints and bounding box labels, with 96 frames per video.

\subsection{Annotation Pipeline}

Most of the related work only employs HRNet \cite{wang2020deep} to estimate the human pose keypoints and present suboptimal performance, which is attributed to the following limitations: 1) The videos depict complex diving actions of the athletes, and HRNet struggles to estimate the precise human pose, due to the poor image quality resulting from high-speed motions and extreme body distortions. 2) The presence of both athletes and spectators within the frame often results in the model incorrectly estimating the pose of the audience members. To address these limitations, we implement a new annotation pipeline that integrates both manual and automatic annotation methodologies. The details are presented as follows:

\textbf{\emph{Collection of manual annotated labels}.} To address the issue of blurred RGB images, we initially implement a series of data preprocessing steps to enhance the quality of the data and then manually annotate the skeletal keypoints. Specifically, we perform the cleaning of the dataset by removing synchronized diving videos and retaining 2,303 individual diving videos. We extract up to 10 videos per action number, based on 48 distinct diving action numbers, resulting in a total of 367 videos. The specific distribution of diving action numbers is provided in the supplementary material. We then extract the airborne movements of the divers across these videos, manually annotating each frame to construct the annotated dataset, which consists of 12,722 manually labeled images. These images encompass various poses, such as pike, tuck, and twist, with a resolution of 455 × 256 pixels for each image.

We utilize the LabelMe \cite{russell2008labelme} tool for annotation and verification. The annotations adhere to the MPII data standard~\cite{andriluka20142d}, encompassing the bounding box of the athlete and 16 pose keypoints. The bounding box is rectangular, with the top-left and bottom-right coordinates clearly annotated. The keypoints are represented as points, with the coordinates of their centers marked from 0 to 15. After the first round of annotation, the labels from one annotator are reviewed and refined by another annotator to ensure that each image contains annotations for both the bounding box and all 16 keypoints. Additional details about the keypoints and manual annotation process are provided in the supplementary material.

\begin{figure*}[t]
  \centering
  \includegraphics[width=0.85\linewidth]{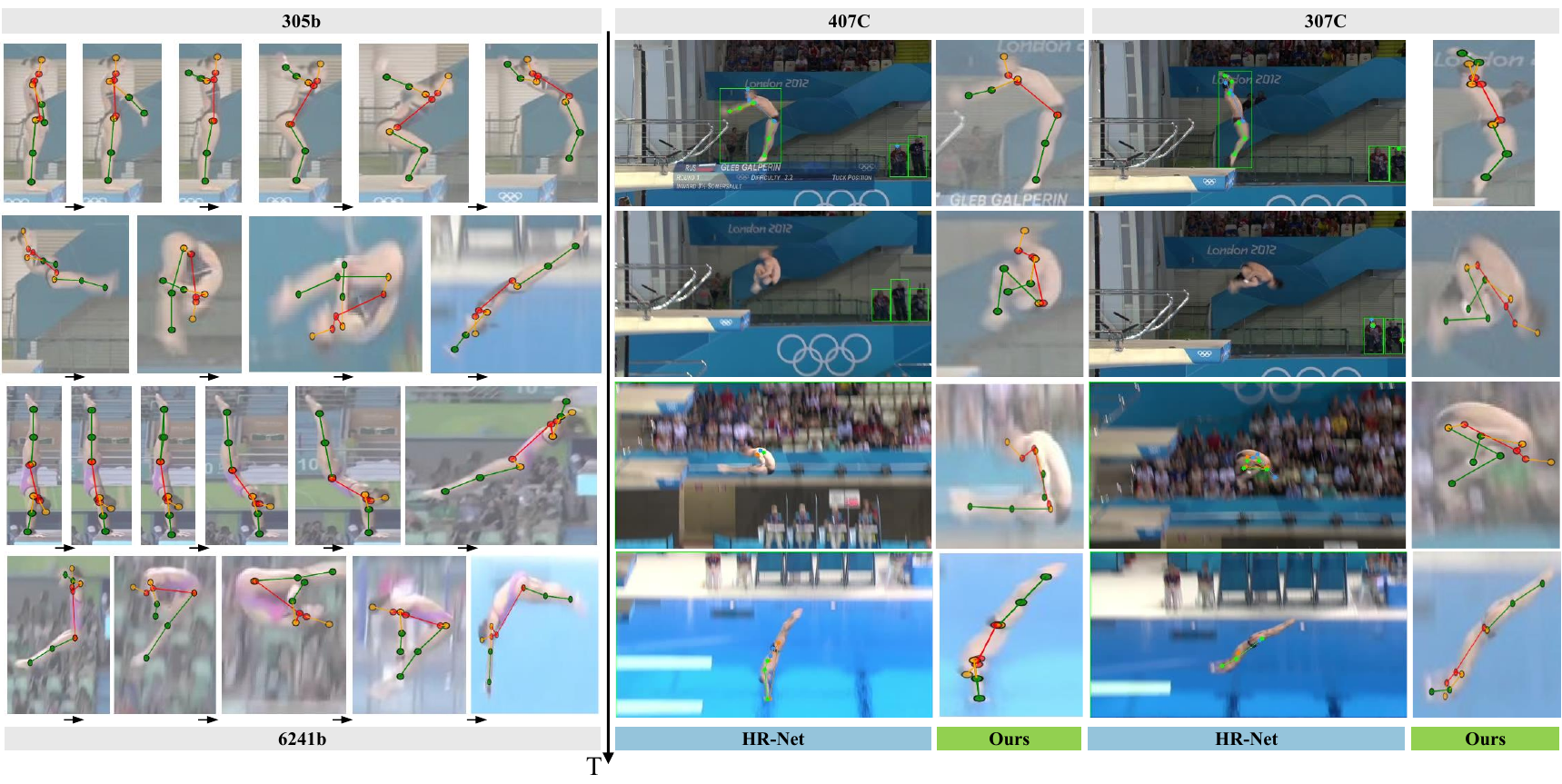}
  \vspace{-3mm}
  \caption{\ar{Visualization results of the FineDiving-Pose dataset. The left panel presents detailed 2D skeletal annotations covering the complete action sequence, including takeoff, flight, and entry. While the right panel presents a comparative visualization of actions `407C' and `307C' using the original HRNet and our proposed method. Compared with the standard HRNet approach, our method demonstrates accuracy in handling extreme body postures.}}
  \vspace{-5mm}
  \label{fig4}
\end{figure*}

\textbf{\emph{Collection of automatic annotated labels}.} The 12,722 annotated images serve as the training set to finetune the HRNet. This process builds an athlete pose estimation network that achieves multi-scale fusion by exchanging information across parallel multi-resolution sub-networks. The network is responsible for resolving blurry images and extreme body distortions, thereby enabling more accurate estimation of highly contorted human poses. Prior to performing pose recognition, we propose to detect human targets in the image, locating all bounding boxes, and then using the pose estimator to detect keypoints within those bounding boxes. For target tracking, we opt for a simple yet effective method: the nearest neighbor algorithm, rather than using deep learning-based approaches. The primary rationale behind this choice is that in extreme pose scenarios, such as diving, the variation in human pose distribution is significant. Additionally, interference from spectators, referees, and other normal human poses complicates the task of tracking divers effectively using deep learning methods.

Specifically, given the athlete coordinates in the $k$-th frame are $p_k=(x_1,y_1,x_2,y_2)_k$, and the potential human coordinates in the next frame are $\left(x_1^i,y_1^i,x_2^i,y_2^i\right)_{k+1}$, where $i$ represents the potential human target, then we calculate the distance between the $i$-th  detected bounding box in the $k+1$-th frame and other confirmed bounding boxes as follows:
\begin{equation}
    dis_{k+1}^i=\sqrt{\sum_{j=1}^2\left(\left(x_j^i-x_j\right)^2+\left(y_j^i-y_j\right)^2\right)}.
\end{equation}
Afterwards, we determine the final subject based on the confidence score and distance among human bounding boxes. Then, we adopt HRNet to obtain the accurate athlete poses estimation based on the subject's coordinates, as the following:
\begin{equation}
    P^q=\text{HRNet}(\left\{V_{(x_1,y_1,x_2,y_2)_k}^q\right\}_{k=1}^T),
\end{equation}
where $\{V_k\}_{k=1}^{T}$ denotes the cropped images from frame 1 to frame $T$.
Finally, for skeletal frames with incomplete joint sequences, joint interpolation is applied to reconstruct missing information. The interpolation formula is listed as follows:
\begin{equation}
    p_k=p_i\frac{k-i}{j-i}+p_j\frac{j-k}{j-i},
\end{equation}
where $p_i$ and $p_j$ denote the coordinates of the corresponding skeletal nodes, and $i$, $j$ means the $i$-th/$j$-th frame respectively. 

\subsection{Visualization Analysis of Annotated Data}
\ar{The right side of Figure \ref{fig4} illustrates the visualized differences between our introduced data annotations and the skeletal poses annotated by HRNet. We show various actions in the visualizations to demonstrate the effectiveness of our approach. For actions such as `407C' and `307C', we observe that for simpler actions (\emph{e.g.}, takeoff and final entry into the water), HRNet mostly identifies them correctly. However, during extreme poses, when there are multiple normal poses in the frame (e.g., referees and spectators), the model tends to focus more on the normally posed individuals. Furthermore, due to the twisted nature of human motion in extreme poses, the high-speed blurry video conditions further degrade the model's performance. In contrast, our proposed annotated method (as shown in the second column of  `407C') can effectively identify the athlete's pose even under high-speed blur. \arr{Meanwhile, the left panel of Figure~\ref{fig4} shows skeletal annotations across consecutive frames, we can see even under fast rotational movements (\emph{e.g.}, 305b, 6241b), our pose estimator maintains consistent tracking of body joints across frames.}}

\section{EXPERIMENT}

In this section, we detail the experimental settings and present the results of our evaluations. We evaluate our proposed approach on two widely-adopted AQA datasets, \emph{i.e.,} \emph{FineDiving} \cite{xu2022finediving} and \emph{MTL-AQA} \cite{parmar2019and}.

\subsection{Experiment Settings}
\textbf{FineDiving Dataset} \cite{xu2022finediving} comprises 3000 video samples, encompassing 52 dive numbers, 29 sub-action numbers, and 23 difficulty degree types. It provides exhaustive annotations, encompassing dive numbers (DN), sub-action numbers, coarse- and fine-grained temporal boundaries, as well as action scores-features. Here, DN refers to the action type labels, indicative of the specific sequence and categories of movements during a dive. Take DN `407C' as an example, the first digit 4, represents an inward dive, the third digit 7, denotes the number of half somersaults to be completed, and `C' signifies the tuck position. In this paper, the dataset is divided into 2,250 training samples and 750 test samples.

\textbf{MTL-AQA Dataset} \cite{parmar2019and} contains 1,412 fine-grained samples from 16 different events, captured from various perspectives. It provides different types of annotations to support research on multiple tasks, including action quality assessment, action recognition, and commentary generation. Additionally, it includes original score annotations from seven judges and the degree of difficulty (DD) for each action. In this paper, the dataset is divided into 1,059 training samples and 353 test samples.

\textbf{Evaluation Metrics.}
Following~\cite{pan2019action},~\cite{yu2021group},~\cite{xu2024fineparser}, we use Spearman's rank correlation coefficient (SRCC) to quantify the rank correlation between the true scores and predicted scores. A higher SRCC indicates a stronger scoring capability of the model.
The computation of SRCC can be expressed as follows:
\begin{equation}
    \rho=\frac{\sum_{i}(r_{i}-\bar{r})(\hat{r}_{i}-\bar{\hat{r}})}{\sqrt{\sum_{i}(r_{i}-\bar{r})^{2}\sum_{i}(\hat{r}_{i}-\bar{\hat{r}})^{2}}},
\end{equation}
where $r$ and $\hat{r}$ represent the ranking for each sample of two series respectively.
We also employ Relative L2-distance ($R_{\mathcal{L}_{2}}$) as an evaluation metric. 
$R_{\mathcal{L}_{2}}$ can be defined as follows:
\begin{equation}
    R_{\mathcal{L}_{2}}=\frac{1}{N}\sum_{i=1}^N (\frac{|y_i-\hat{y}_i|}{y_{max}-y_{min}}).
\end{equation}
Here, \(y_i\) and \(\hat{y}_i\) represent the ground truth and predicted scores for the \textit{i}-th sample, respectively. Lower $R_{\mathcal{L}_{2}}$ indicates better performance of the approach.
The computation of SRCC and $R_{\mathcal{L}_{2}}$ metrics in detail can be found in the paper \cite{yu2021group}.

Afterwards, we obtain the one-dimensional boundary box for each stage $[\hat{t}_{k-1},\hat{t}_{k}]_{k=1}^{\mathbb K-1}$. Assuming the ground truth boundary box denoted as $[t_{k-1},t_{k}]_{k=1}^{\mathbb K-1}$, we calculate the average Intersection over Union (IoU) between the predicted and ground truth boundary boxes. A prediction is considered correct if the IoU exceeds a certain threshold $d$. Thus, AIoU@$d$ is calculated as follows:
\begin{equation}
    \mathrm{AIoU@}d=\frac{1}{N}\sum_{i=1}^{N}\mathcal{J}\left(\mathrm{IoU}_{i}\geq d\right),
\end{equation}
where N is the number of samples. If the judgment is true, \(\mathcal{J}=1\), otherwise, \(\mathcal{J}=0\). In this work, we primarily adopt AIoU@0.5 and AIoU@0.75 as the main metrics. 

\ar{\textbf{Baseline Methods.} We compare our method with the following approaches including Pose+DCT~\cite{pirsiavash2014assessing}, C3D-LSTM~\cite{pan2019action}, MUSDL~\cite{tang2020uncertainty}, CoRe~\cite{yu2021group}, TSA~\cite{xu2022finediving}, TSA-Net~\cite{wang2021tsa}, H-GCN~\cite{zhou2023hierarchical}, PECoP~\cite{dadashzadeh2023pecopparameterefficientcontinual}, T2CR~\cite{ke2024two}, TPT~\cite{bai2022action} and MCoRe~\cite{an2024multi}. We adopt the publicly available code repositories to reproduce CoRe~\cite{yu2021group}, H-GCN~\cite{zhou2023hierarchical}, TPT~\cite{bai2022action} and T2CR~\cite{ke2024two} according to the parameter settings in the original papers, and we denote these results with an asterisk (*). For other methods, we cite the results reported in the original papers.}

\textbf{Implementation Details.} We implemented our proposed method using PyTorch deep learning framework. The I3D model \cite{carreira2017quo} pre-trained on the Kinetics dataset was utilized as the dynamic visual encoder.  This encoder comprises the spatiotemporal feature extraction module and the mixed convolution module.  The HD-GCN model~\cite{lee2023hierarchically} was adopted as the base model for the hierarchical skeletal encoder.  In the multimodal fusion Transformer, the number of attention heads was set to 2.  Additionally, the static visual encoder uses ResNetY \cite{he2016deep} as the base model.  The stage-wise decoder uses a multi-head attention network to calculate differences between various stages, with 4 heads used during training.  The initial learning rate for the aforementioned models was set to $5e - 4$, and the Adam optimizer was used with zero weight decay.  Our model was trained with a batch size of 8 for 200 epochs.  \arr{Compared to existing methods~\cite{yu2021group,xu2022finediving} that typically use a batch size of 2, we adopt a larger batch size to achieve faster training speed.} Following \cite{xu2022finediving}, we extracted 96 frames from each video in the FineDiving dataset and divided them into 9 clips in I3D.  In the experiments on the MTL-AQA dataset, we extracted 103 frames per video clip following~\cite{yu2021group}, and segmented them into 10 overlapping segments, each containing 16 consecutive frames.  During inference, we sampled 10 exemplar videos for each query video, and aggregated their scores using a multi-sample voting strategy. We set the number of stage transitions \(\mathbb K\) as 3 according to the different stages of diving (\emph{i.e.,} take-off, flying and entry).

\subsection{Quantitative Results}

\begin{table}[t]
  \centering
  \caption{\ar{comparison results with the state-of-the-art methods on finediving dataset. best results are in bold. here w/dn and w/o dn refer to with and without using dive numbers.} }
  \vspace{-2mm}
  \ar{
  \begin{tabular}{ccccc}
    \toprule
    \multirow{2}{*}{Methods (w/o DN)} & \multirow{2}{*}{Year} & \multirow{2}{*}{$\rho$} &\multirow{2}{*}{$ R_{\mathcal{L}_{2}}$($\times$100)} & AIoU\\
    
    &&&&0.5/0.75 \\
    \midrule
    USDL~\cite{tang2020uncertainty} & 2020 & 0.8302 & 0.592 & - \\
    MUSDL~\cite{tang2020uncertainty} & 2020 & 0.8427& 0.573 & - \\
    CoRe~\cite{yu2021group} & 2021 & 0.8631 & 0.556 & - \\
    TSA~\cite{xu2022finediving} & 2022 & \underline{0.8925} & 0.478 & 80.71 / 30.17 \\
    MCoRe~\cite{an2024multi} & 2024 & 0.8897 & \underline{0.461} & 89.42 / 74.46 \\
    \midrule
    \textbf{Ours} & 2024 & \textbf{0.9041} & \textbf{0.411} & \textbf{98.17 / 93.68} \\
    \bottomrule
    \toprule
    \multirow{2}{*}{Methods (w/ DN)} & \multirow{2}{*}{Year} & \multirow{2}{*}{$\rho$} &\multirow{2}{*}{$ R_{\mathcal{L}_{2}}$($\times$100)} & AIoU\\
    
    &&&&0.5/0.75 \\
    \midrule
    USDL~\cite{tang2020uncertainty} & 2020 & 0.8504 & 0.583 & - \\
    MUSDL~\cite{tang2020uncertainty} & 2020 & 0.8978 & 0.370 & - \\
    CoRe~\cite{yu2021group} & 2021 & 0.9061 & 0.361 & - \\
    TSA~\cite{xu2022finediving} & 2022 & 0.9203 & 0.342 & 82.51 / 34.31 \\
    PECoP~\cite{dadashzadeh2023pecopparameterefficientcontinual} & 2023 & \underline{0.9315} & - & - / - \\
    MCoRe~\cite{an2024multi} & 2024 & 0.9232 & 0.326 & 98.26 / 79.17 \\
    T2CR~\cite{ke2024two} & 2024 & 0.9275 & 0.330 & - / - \\
    \midrule
    \textbf{Ours} & 2024 & \textbf{0.9383} & \textbf{0.269} & \textbf{99.07 / 97.96} \\
    \bottomrule
  \end{tabular}
  }
  \label{finediving}
  \vspace{-1mm}
\end{table}

\ar{We compare our proposed method against other AQA methods on FineDiving \cite{xu2022finediving} and MTL-AQA \cite{parmar2019and} datasets. The results for FineDiving dataset are shown in Table~\ref{finediving}, while those for the MTL-AQA dataset are shown in Table~\ref{mtl-aqa}. We can clearly observe that our method significantly improved the SRCC  and the $R_{\mathcal{L}_{2}}$ compared to all other methods on FineDiving. Specifically, our method achieved an SRCC of 0.9041 and an $R_{\mathcal{L}_{2}}$ of 0.411 without using dive numbers. When dive numbers are incorporated, the SRCC improves to 0.9383 and the $R_{\mathcal{L}_{2}}$ reduces to 0.269. Unlike TSA \cite{xu2022finediving}, which uses coarse-grained I3D to segment sub-actions, we introduced a static visual module and hierarchical skeletal features to capture fine-grained differences between sub-actions. Compared to TSA \cite{xu2022finediving}, our method achieved improvements of 1.16\% and 1.8\% in SRCC under the two conditions, respectively, and improvements of 0.067 and 0.073 in $R_{\mathcal{L}_{2}}$. On the MTL-AQA dataset, we evaluate the contribution with and without the difficulty (DD) labels. As shown in Table \ref{mtl-aqa}, our proposed method significantly outperforms other methods in the (w/o DD) setting, achieving an SRCC of 0.9266 and an $R_{\mathcal{L}_{2}}$ of 0.446. Compared to the classic CoRe \cite{yu2021group} method, our approach improves the SRCC by 0.43\% and the $R_{\mathcal{L}_{2}}$ by 0.02. When DD are incorporated, the degree of difficulty serves as the reference standard for the distribution learning of H-GCN~\cite{zhou2023hierarchical} and TPT~\cite{bai2022action}, allowing it to better calibrate the predicted distribution, while our method still achieve the second best performance. These findings demonstrate that Our approach exhibits stronger generalizability, as the hierarchical skeletal encoder effectively guides the model to learn differences between actions even without DD labels. }

\begin{table}[t]
  \centering
  \caption{\ar{comparison results with the state-of-the-art methods on MTL-AQA Dataset. Best results are in bold and the second best are with underlines. here w/dd and w/o dd refer to with and without degree of difficulty labels. }}
  \ar{
  \begin{tabular}{cccc}
    \toprule
    Approaches (w/o DD) & Year & $\rho$ &$ R_{\mathcal{L}_{2}}$($\times$100)  \\
    \midrule
    Pose+DCT~\cite{pirsiavash2014assessing} & 2014 & 0.2682 & -  \\
    C3D-SVR~\cite{pan2019action} & 2017 & 0.7716& -  \\
    C3D-LSTM~\cite{pan2019action} & 2017 & 0.8489 & -  \\
    USDL~\cite{tang2020uncertainty} & 2020 & 0.9066 & 0.654  \\
    MUSDL~\cite{tang2020uncertainty}  & 2020 & 0.9158 & 0.609  \\
    CoRe*~\cite{yu2021group} & 2021 & \underline{0.9223} & \underline{0.466}  \\
    H-GCN*~\cite{zhou2023hierarchical}  & 2023 & 0.9178 & 0.487  \\
    \midrule
    \textbf{Ours} & 2024 & \textbf{0.9266} & \textbf{0.446}  \\
    \bottomrule
    \toprule
    Approaches (w/ DD) & Year & $\rho$ &$ R_{\mathcal{L}_{2}}$($\times$100) \\
    \midrule
    USDL~\cite{tang2020uncertainty} & 2020 & 0.9231 & 0.468  \\
    MUSDL~\cite{tang2020uncertainty} & 2020 & 0.9273 & 0.451  \\
    CoRe*~\cite{yu2021group} & 2021 & 0.9423 & 0.336  \\
    TSA-Net~\cite{wang2021tsa}  & 2021 & 0.9422 & -  \\
    TPT*~\cite{bai2022action}  & 2022 & \textbf{0.9552} & \textbf{0.2542}  \\
    H-GCN*~\cite{zhou2023hierarchical}  & 2023 & 0.9463 & 0.311  \\
    T2CR*~\cite{ke2024two}  & 2024 & 0.9429 & 0.368  \\
    \midrule
    \textbf{Ours} & 2024 & \underline{0.9516} & \underline{0.278}  \\
    \bottomrule
  \end{tabular}
  }
  \label{mtl-aqa}
  \vspace{-2mm}
\end{table}

\begin{table}[t]
  \centering
  \caption{evaluation on components of multi-scale visual-skeletal encoder in the finediving dataset.}
  \vspace{-2mm}
\begin{tabular}{cccccc}
    \toprule
    Approaches & DVE & SVE & HSE & $\rho$ & $ R_{\mathcal{L}_{2}}$($\times$100) \\
    \midrule
    DVE & $\checkmark$ & $\text{\sffamily x}$ & $\text{\sffamily x}$ & 0.8914 & 0.4784 \\
    SVE & $\text{\sffamily x}$ & $\checkmark$ & $\text{\sffamily x}$ & 0.9173 & 0.3945 \\
    HSE & $\text{\sffamily x}$ & $\text{\sffamily x}$ & $\checkmark$ & 0.6847 & 0.9871 \\
    DVE+HSE & $\checkmark$ & $\text{\sffamily x}$  &$\checkmark$ & 0.9271 & 0.3377 \\
    DVE+SVE & $\checkmark$ & $\checkmark$  &$\text{\sffamily x}$ & 0.9318 & 0.2914 \\
    Ours Full & $\checkmark$ & $\checkmark$ & $\checkmark$ & \textbf{0.9383} & \textbf{0.2691} \\
    \bottomrule
\end{tabular}
  \label{ablation}
   \vspace{-2mm}
\end{table}

\begin{table}[h]
\centering
\caption{Ablation study on the FineDiving dataset.}
\vspace{-1mm}
\ar{
\begin{tabular}{lcc}
    \toprule
    Ablation study & $\rho$ & $R_{\mathcal{L}_{2}}$($\times$100) \\
    \midrule
    \multicolumn{3}{l}{\textbf{(a) Hierarchical Skeletal Encoder}} \\
    w/ $H_0$                & 0.9316  & 0.2867  \\
    w/ $H_1$                & 0.9310  & 0.2891  \\
    w/ $H_2$                & 0.9324  & 0.2814  \\
    w/ $H_0+H_1$            & 0.9334  & 0.2703  \\
    w/o $EdgeConv$          & 0.9337  & 0.2665  \\
    Ours                    & \textbf{0.9383} & \textbf{0.2691} \\
    \midrule
    \multicolumn{3}{l}{\textbf{(b) Number of Stages $\mathbb{K}$}} \\
    1                       & 0.9247  & 0.3522  \\
    2                       & 0.9328  & 0.2884  \\
    2*                      & 0.9316  & 0.2944  \\
    3                       & \textbf{0.9383} & \textbf{0.2691} \\
    \midrule
    \multicolumn{3}{l}{\textbf{(c) Number of Temporal Samples $T$}} \\
    3                       & 0.9353  & 0.2469  \\
    5                       & \textbf{0.9383} & \textbf{0.2691} \\
    8                       & 0.9330  & 0.2552  \\
    10                      & 0.9357  & 0.2455  \\
    \midrule
    \multicolumn{3}{l}{\textbf{(d) Fusion Methods for MFM}} \\
    Addition                & 0.9341  & 0.2874  \\
    Dot Product             & 0.9226  & 0.3189  \\
    Weighted Addition (Ours)& \textbf{0.9383} & \textbf{0.2691} \\
    \midrule
    \multicolumn{3}{l}{\textbf{(e) Contrastive Regression Loss}} \\
    InfoNCE Loss\cite{oord2018representation} & 0.9142 & 0.3526 \\
    Triplet Loss \cite{schultz2003learning}   & 0.9302 & 0.2962 \\
    Ours                    & \textbf{0.9383} & \textbf{0.2691} \\
    \bottomrule
    \vspace{-3mm}
\end{tabular}
}

\label{tab:ablation_finediving}
\end{table}

\begin{table}[h]
\centering
\caption{Evaluation on the network of procedure segmentation module}
\vspace{-1mm}
\begin{tabular}{cccc}
    \toprule
    
    \multirow{2}{*}{ Approaches } & \multirow{2}{*}{$\rho$} &\multirow{2}{*}{$ R_{\mathcal{L}_{2}}$($\times$100)} & AIoU\\
    
    &&&0.5/0.75 \\
    
    \midrule
    FC & 0.9230  & 0.3178 & 88.17 / 71.51\\
    TCN \cite{bai2018empirical}  & 0.9136  & 0.3725 & 83.17 / 34.99\\
    ASFormer \cite{yi2021asformer} & 0.9357  & 0.2672 & 99.13 / 89.71\\
    Ours & \textbf{0.9383} & \textbf{0.2691} & \textbf{99.07} / \textbf{97.96}\\
    \bottomrule
\end{tabular}
\label{Procedure Segmentation Module}
\vspace{-2mm}
\end{table}

\subsection{Ablation Study}
We conduct a series of ablation experiments on our proposed method using the FineDiving dataset.

\textbf{Multi-scale Visual-Skeletal Encoder.} We analyze the effectiveness of each component in our multi-scale visual-skeletal model, including the dynamic visual encoder (DVE), static visual encoder (SVE), and hierarchical skeletal encoder (HSE). In the experiments, the number of stages is set as $\mathbb K=3$, the number of layers in the Transformer decoder is $4$, and the number of attention heads is $2$. We evaluate each component in the Multi-scale Visual-Skeleton Encoder (MVSE) and present the results in Table ~\ref{ablation}. The results demonstrate that each component in the MVSE plays a crucial role. Specifically, incorporating the HSE enables the model to capture subtle differences between sub-actions, achieving a 3.57\% improvement in SRCC compared to the baseline. Additionally, utilizing the HSE alone results in inferior performance, with an SRCC of only 0.6847. It is attributed to the absence of some crucial visual information, such as splash size, which is vital for accurately evaluating the quality of the diving action.

\begin{figure*}[t]
  \centering
  \includegraphics[width=0.9\linewidth]{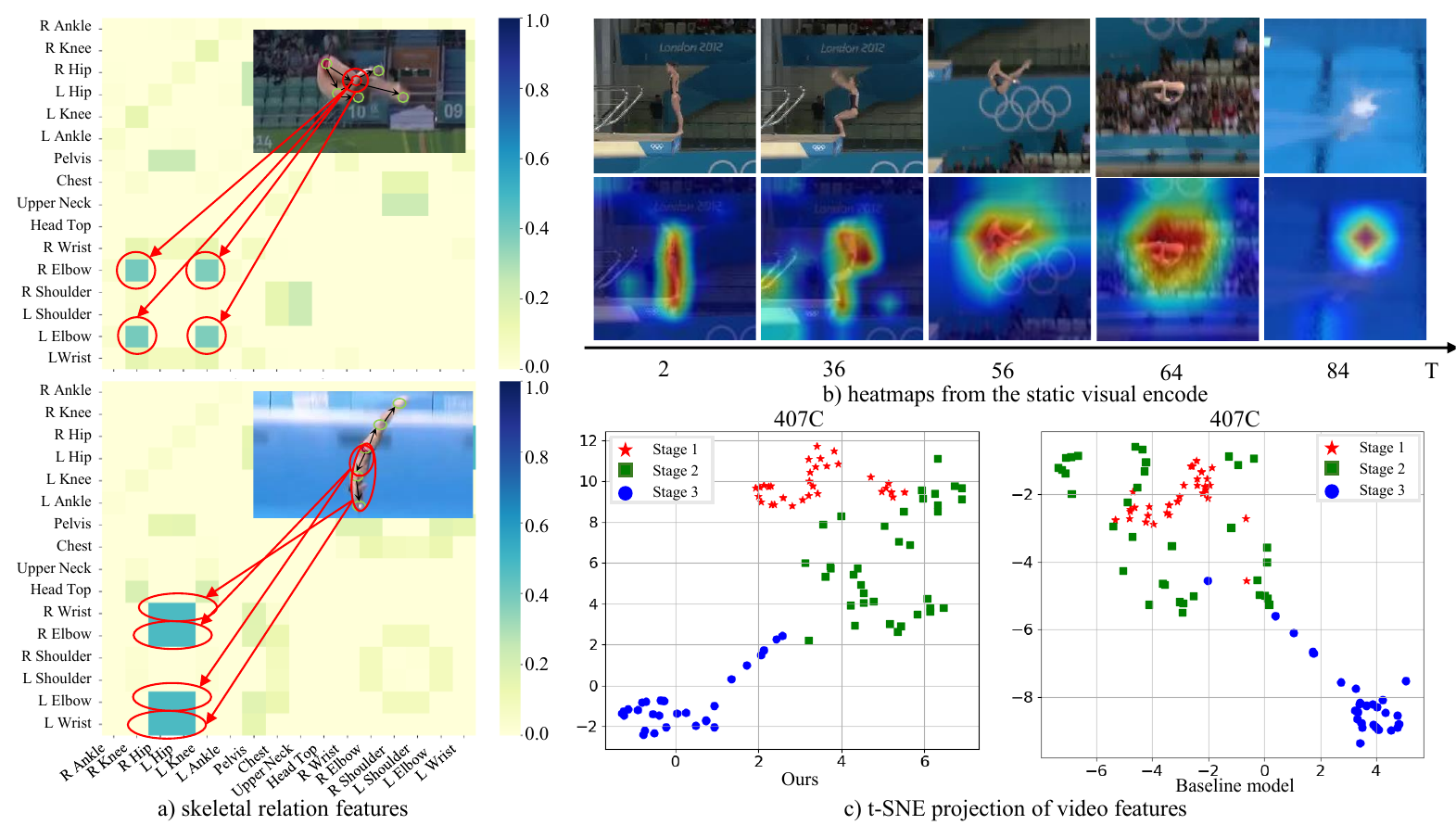}
  \vspace{-2mm}
  \caption{\ar{Visualization of skeletal relation features, heatmaps from the static visual encoder, and t-SNE projection of video features. a) The left panel illustrates the learned skeletal relational features for two distinct actions: the tuck position and the water entry. \arr{b) The top-right panel shows the heatmap visualization generated by the Grad-CAM~\cite{selvaraju2017grad} tool from the static visual encoder, where the horizontal axis represents the frame sequence of the video.} c) The bottom-right panel presents the t-SNE visualization of the video features, with different sub-action stages labeled using distinct colors and shapes to highlight semantic separability in the feature space.}}
  \vspace{-3mm}
  \label{fig8}
\end{figure*}

\textbf{Hierarchical Skeletal Encoder.} We primarily analyze the impact of different levels of skeletal joints on the model’s ability. Specifically, we separately input the sub-action features learned by three different skeletal topologies into the multimodal fusion model to evaluate the effectiveness of each level. We also assess the impact of adding EdgeConv for semantic edge modeling. As shown in Table \ref{tab:ablation_finediving} (a), the experimental results indicate that different skeletal topologies have varying effects on guiding the model to learn sub-actions. For example, $H_2$, which is employed to capture the coordination level of the athlete's body movements, significantly enhances the model's assessment of movement quality compared to $H_0$ and $H_1$. This is reflected by an SRCC improvement of 0.14\% over the other two levels. Moreover, we explored the impact of semantic edges on improving the performance. Even the combination of $H_0$ and $H_1$ brings improvements, our full model including three levles human pose still achieves the the best. By removing the EdgeConv, the result drops by 4.6\%. This finding suggests that incorporating fully connected semantic information can enhance the generalization ability.

\textbf{Procedure Segmentation Module.} In the procedure segmentation module, we focus on the impact of the number of divided actions or distinct sub-action stages. Taking diving as an example, an athlete’s action can be segmented into three stages: takeoff, flight, and entry. The goal of this segmentation is to enable reasonable and accurate finer-grained scoring. Specifically, we explore the results under different segmentation scenarios. When  $\mathbb K=2$,  the model divides the action into two stages using a single segmentation point. This segmentation point either separates takeoff or entry (denoted with *) from the whole movement. As shown in Table \ref{tab:ablation_finediving} (b), segmenting the stages during the diving process facilitates the model’s learning of finer-grained sub-actions. Compared to  $\mathbb K=1$ or $2$, our method achieves up to a maximum of 1.36\% in SRCC. Additionally, we conduct ablation experiments on the various backbones of the procedure segmentation network. We compare three different action segmentation models, \emph{i.e.,} ASFormer \cite{yi2021asformer}, TCN \cite{bai2018empirical}, and using only a fully connected layer. As shown in Table \ref{Procedure Segmentation Module}, compared to other methods, our proposed approach is based on Bi-GRU, which utilizes gated units to control the flow and retention of temporal information, thereby enhancing the model’s ability to remember long-term dependencies and achieving the best performance. Additionally, in order to evaluate the impact of the temporal sampling size  $T$  on stage segmentation, we conducted a series of experiments, with the results being summarized in Table \ref{tab:ablation_finediving} (c). We can observe that when $T$ is too small, it results in insufficient semantic information, negatively affecting model performance, when $T$ is too large, the performance reaches saturation, while increasing the complexity of the model. When $T$ equals 5, our proposed model achieves the best performance and hence obtain a good trade-off. 

\textbf{Discussion on Multi-Modal Fusion Methods.} We evaluate the performance of various fusion methods for combining visual and skeletal features. Specifically, we test three approaches: Element-wise addition, Dot product, and our adopted Weighted addition. The experimental results are shown in Table \ref{tab:ablation_finediving} (d). Compared to simple addition or multiplication of multimodal attention maps, our proposed method incorporates learnable fusion parameters to perform a weighted summation of visual and skeletal attention maps.  Unlike the dot product, which often leads to attention sparsity, the weighted addition approach demonstrates superior representational capability, achieving a 1.57\% improvement in SRCC results.

\begin{figure*}[t]
  \centering
  \includegraphics[width=0.85\linewidth]{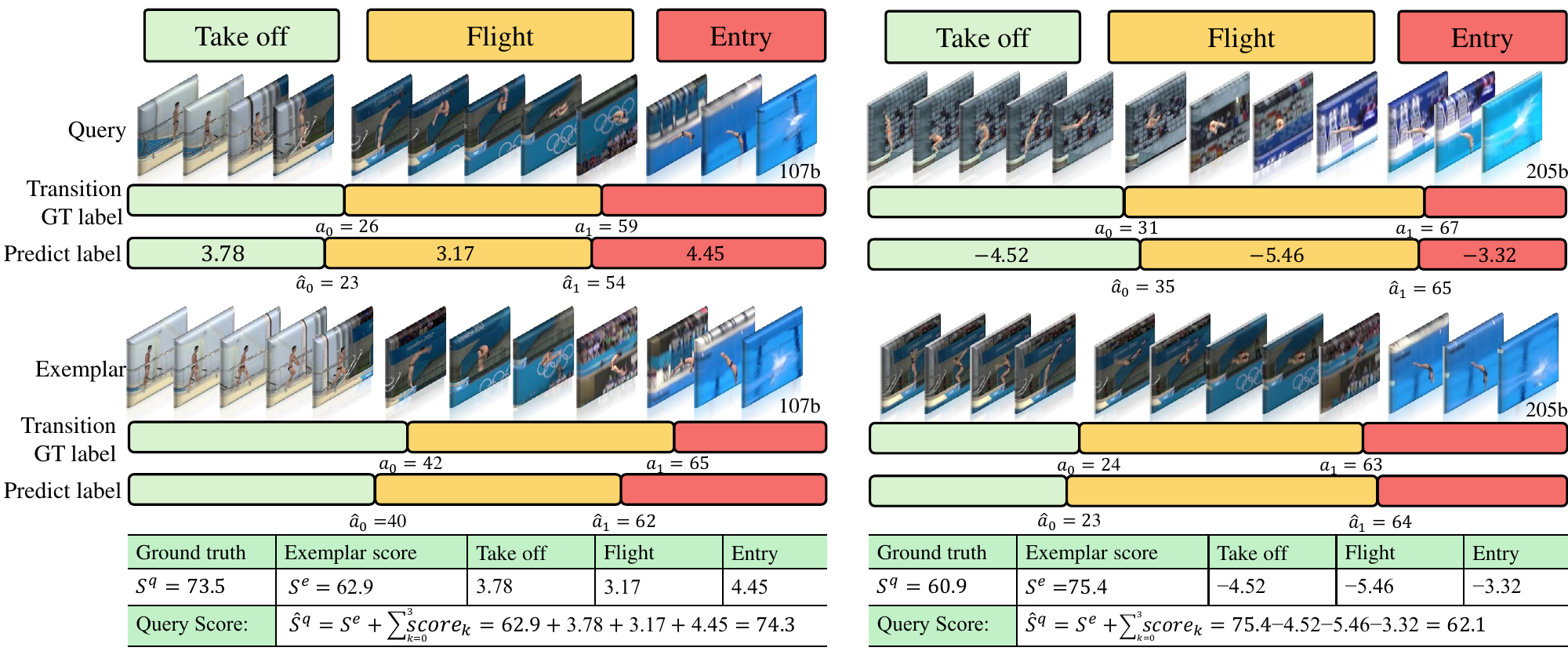}
  \vspace{-1mm}
  \caption{Visualization of predicted results. The input consists of a query video and an exemplar video. The videos are segmented into sub-actions, enabling stage predictions of relative scores for each sub-action. Subsequently, these relative scores are aggregated with the exemplar scores to compute the final score. } 
  \vspace{-3mm}
  \label{fig7}
\end{figure*}

\textbf{Stage-wise Contrastive Loss.} In the multi-stage contrastive regression module, we evaluate the effectiveness of our proposed stage-wise contrastive loss by comparing it with two widely used contrastive losses: InfoNCE Loss \cite{oord2018representation} and Triplet Loss \cite{schultz2003learning}. The experimental results are presented in Table \ref{tab:ablation_finediving} (e). Compared to InfoNCE, which relies on a large number of negative samples to ensure effective learning, our approach leverages both intra-video and inter-video negative samples from multiple sources. This design enables our method to maintain effective learning with a significantly reduced number of contrastive samples. Compared to  InfoNCE loss, our method achieves an improvement of 2.41\% in SRCC.

\textbf{Stage-wise Contrastive Loss.} In the multi-stage contrastive regression module, we evaluate the effectiveness of our proposed stage-wise contrastive loss by comparing it with two widely used contrastive losses: InfoNCE Loss \cite{oord2018representation} and Triplet Loss \cite{schultz2003learning}. The experimental results are presented in Table \ref{tab:ablation_finediving} (e). Compared to InfoNCE, which relies on a large number of negative samples to ensure effective learning, our approach leverages both intra-video and inter-video negative samples from multiple sources. This design enables our method to maintain effective learning with a significantly reduced number of contrastive samples. Compared to  InfoNCE loss, our method achieves an improvement of 2.41\% in SRCC.

\begin{table}[t]
    \centering
    \caption{\ar{Computation cost and performance comparison on the MTL-AQA dataset.}}
    \vspace{-2mm}
    \label{tab:computation cost}
    \ar{
    \begin{tabular}{l c c c}
        \toprule
        \textbf{Model} & \textbf{GFLOPs} & \textbf{Params} & \textbf{$\rho$}\\
        \midrule
        TSA~\cite{xu2022finediving} & $1008.7$ & $12.6$M & 0.9470 \\
        MCORE~\cite{an2024multi} & $45.5$ & $3.1$M &  0.9351\\
        T2CR~\cite{ke2024two} & $252.3$ & $12.2$M & 0.9430\\
        TPT~\cite{bai2022action} & $557.64$ & $16.89$M & \textbf{0.9552}\\
        Our model w/o Skeletal Encoder & $137.64$ & $16.62$M & -\\
        Our Full model & $138.1$ & $16.8$M & \underline{0.9516}\\
        \bottomrule
    \end{tabular}
    }
    \vspace{-2mm}
\end{table}

\ar{\textbf{Computation Cost Analysis.} As shown in Table \ref{tab:computation cost}, our method achieves lower GFLOPs and fewer parameters than state-of-the-art models such as TPT and TSA, while maintaining competitive performance. This efficiency gain mainly stems from a lightweight visual backbone and an optimized skeletal encoder design.}

\subsection{Visualization Results}
\ar{We mainly visualize the spatial relationships of skeletal features and present the final segmentation results. we present the spatial relation graphs of the skeletal features, as shown in Figure~\ref{fig8}. In Figure~\ref{fig8} a), the athlete is performing a tuck position during the somersault phase of the dive, darker colors indicate higher attention weights assigned by the model to specific joint pairs. At this stage, the key focus is on minimizing the distances between the elbows and knees to achieve a compact and standard tuck posture. In the bottom top-left sub-figure, the athlete is in the entry phase of the dive. Here, the athlete is entering the water, where body verticality is crucial, particularly the extension of the hips, elbows, and wrists. The corresponding graph shows that the model attends more to the spatial coordination among these joints during this phase. \arr{In Figure~\ref{fig8} b), we adopt the Grad-CAM~\cite{selvaraju2017grad} tool to generate visualized attention heatmaps from the static encoder, which effectively captures static human appearance features across the frame sequence.} Finally, In Figure~\ref{fig8} c), we employ the t-SNE~\cite{tsne} method to perform dimensionality reduction and visualization of the static visual features, in order to validate the effectiveness of the proposed multi-stage contrastive learning approach. Compared to the baseline ResNetY model~\cite{radosavovic2020designing}, the features extracted by our model at various stages exhibit clearer class separability in the reduced feature space, demonstrating enhanced discriminative power enabled by our contrastive learning strategy.}

As illustrated in Figure \ref{fig7}, our proposed method begins by segmenting both the query and exemplar videos into sub-actions. In terms of diving, sub-actions are typically divided into three stages: `Takeoff', `Flight' and `Entry'. We can see from the figure that our proposed procedure segmentation module can accurately predict stage transition labels and effectively identify transitions between sub-actions. Subsequently, for each sub-action, our proposed multi-stage contrastive regression module can predict more accurate scores by comparing the query video with the exemplar video. For example, in the ``Entry” stage of `107b' in Figure \ref{fig7}, the splash size in the query video is significantly smaller than that in the exemplar video. As a result, the relative score of the query video’s ``Entry” stage is noticeably higher than that of the exemplar video. Thus, the relative score for the “Entry” stage of the query video is notably higher than that of the exemplar video. Finally, the relative scores of each sub-action are summed with the ground truth score $S^e$ of the exemplar video to compute the predicted score $\hat{S}^q$ for the query video. 

\section{CONCLUSION}
In this paper, we presented a hierarchically pose-guided multi-stage framework for action quality assessment, incorporating a multi-scale visual–skeletal encoder for modality-specific features and a procedure segmentation network for sub-action modeling. A multi-stage contrastive regression was employed to learn discriminative representations, achieving superior results on two challenging AQA benchmarks and our newly annotated FineDiving-Pose dataset. For future work, we plan to explore 3D pose modeling and athlete-related contextual cues within multimodal large language models to enhance tactical and health analysis. 

\bibliographystyle{IEEEtran}
\bibliography{new_refs}

\begin{thebibliography}{10}
\providecommand{\url}[1]{#1}
\csname url@samestyle\endcsname
\providecommand{\newblock}{\relax}
\providecommand{\bibinfo}[2]{#2}
\providecommand{\BIBentrySTDinterwordspacing}{\spaceskip=0pt\relax}
\providecommand{\BIBentryALTinterwordstretchfactor}{4}
\providecommand{\BIBentryALTinterwordspacing}{\spaceskip=\fontdimen2\font plus
\BIBentryALTinterwordstretchfactor\fontdimen3\font minus \fontdimen4\font\relax}
\providecommand{\BIBforeignlanguage}[2]{{%
\expandafter\ifx\csname l@#1\endcsname\relax
\typeout{** WARNING: IEEEtran.bst: No hyphenation pattern has been}%
\typeout{** loaded for the language `#1'. Using the pattern for}%
\typeout{** the default language instead.}%
\else
\language=\csname l@#1\endcsname
\fi
#2}}
\providecommand{\BIBdecl}{\relax}
\BIBdecl

\bibitem{pan2019action}
J.~Pan, J.~Gao, and W.~Zheng, ``Action assessment by joint relation graphs,'' in \emph{Proc. IEEE Conf. Comput. Vis. Pattern Recognit.}, 2019, pp. 6331--6340.

\bibitem{yu2021group}
X.~Yu, Y.~Rao, W.~Zhao, J.~Lu, and J.~Zhou, ``Group-aware contrastive regression for action quality assessment,'' in \emph{Proc. IEEE Int. Conf. Comput. Vis.}, 2021, pp. 7919--7928.

\bibitem{pirsiavash2014assessing}
H.~Pirsiavash, C.~Vondrick, and A.~Torralba, ``Assessing the quality of actions,'' in \emph{Proc. Eur. Conf. Comput. Vis.}, 2014, pp. 556--571.

\bibitem{tang2020uncertainty}
Y.~Tang, Z.~Ni, J.~Zhou, D.~Zhang, J.~Lu, Y.~Wu, and J.~Zhou, ``Uncertainty-aware score distribution learning for action quality assessment,'' in \emph{Proc. IEEE Conf. Comput. Vis. Pattern Recognit.}, 2020, pp. 9839--9848.

\bibitem{bai2022action}
Y.~Bai, D.~Zhou, S.~Zhang, J.~Wang, E.~Ding, Y.~Guan, Y.~Long, and J.~Wang, ``Action quality assessment with temporal parsing transformer,'' in \emph{Proc. Eur. Conf. Comput. Vis.}, 2022, pp. 422--438.

\bibitem{parmar2019and}
P.~Parmar and B.~T. Morris, ``What and how well you performed? a multitask learning approach to action quality assessment,'' in \emph{Proc. IEEE Conf. Comput. Vis. Pattern Recognit.}, 2019, pp. 304--313.

\bibitem{shao2020intra}
D.~Shao, Y.~Zhao, B.~Dai, and D.~Lin, ``Intra- and inter-action understanding via temporal action parsing,'' in \emph{Proc. IEEE Conf. Comput. Vis. Pattern Recognit.}, 2020, pp. 730--739.

\bibitem{zia2018video}
A.~Zia, Y.~Sharma, V.~Bettadapura, E.~L. Sarin, and I.~Essa, ``Video- and accelerometer-based motion analysis for automated surgical skills assessment,'' \emph{Int. J. Comput. Assist. Radiol. Surg.}, vol.~13, no.~3, pp. 443--455, 2018.

\bibitem{parmar2019action}
P.~Parmar and B.~Morris, ``Action quality assessment across multiple actions,'' in \emph{Proc. IEEE Winter Conf. Appl. Comput. Vis.}, 2019, pp. 1468--1476.

\bibitem{xu2019learning}
C.~Xu, Y.~Fu, B.~Zhang, Z.~Chen, Y.~Jiang, and X.~Xue, ``Learning to score figure skating sport videos,'' \emph{IEEE Trans. Circuits Syst. Video Technol.}, vol.~30, no.~12, pp. 4578--4590, 2019.

\bibitem{zhang2015relative}
Q.~Zhang and B.~Li, ``Relative hidden markov models for video-based evaluation of motion skills in surgical training,'' \emph{IEEE Trans. Pattern Anal. Mach. Intell.}, vol.~37, no.~6, pp. 1206--1218, 2015.

\bibitem{sharma2014video}
Y.~Sharma, V.~Bettadapura, N.~Hammerla, S.~Mellor, R.~McNaney, P.~Olivier, S.~Deshmukh, A.~McCaskie, and I.~Essa, ``Video-based assessment of osats using sequential motion textures,'' in \emph{Proc. 5th Workshop Modeling Monitoring Comput. Assisted Interventions}, 2014, pp. 31--40.

\bibitem{zia2015automated}
A.~Zia, Y.~Sharma, V.~Bettadapura, E.~L. Sarin, M.~A. Clements, and I.~Essa, ``Automated assessment of surgical skills using frequency analysis,'' in \emph{Proc. Int. Conf. Med. Image Comput. Comput.Assist. Intervent.}, 2015, pp. 430--438.

\bibitem{du2023semantics}
Z.~Du, D.~He, X.~Wang, and Q.~Wang, ``Learning semantics-guided representations for scoring figure skating,'' \emph{IEEE Transactions on Multimedia}, vol.~26, pp. 4987--4997, 2023.

\bibitem{10737230}
K.~Gedamu, Y.~Ji, Y.~Yang, J.~Shao, and H.~Tao~Shen, ``Visual-semantic alignment temporal parsing for action quality assessment,'' \emph{IEEE Transactions on Circuits and Systems for Video Technology}, vol.~35, no.~3, pp. 2436--2449, 2025.

\bibitem{xu2024vision}
H.~Xu, X.~Ke, Y.~Li, R.~Xu, H.~Wu, X.~Lin, and W.~Guo, ``Vision-language action knowledge learning for semantic-aware action quality assessment,'' in \emph{Proc. Eur. Conf. Comput. Vis.}\hskip 1em plus 0.5em minus 0.4em\relax Cham: Springer Nature Switzerland, 2024, pp. 423--440.

\bibitem{majeedi2024rica2rubricinformedcalibratedassessment}
A.~Majeedi, V.~R. Gajjala, S.~S. S.~N. GNVV, and Y.~Li, ``Rica2: Rubric-informed, calibrated assessment of actions,'' 2024.

\bibitem{xu2022finediving}
J.~Xu, Y.~Rao, X.~Yu, G.~Chen, J.~Zhou, and J.~Lu, ``Finediving: A fine-grained dataset for procedure-aware action quality assessment,'' in \emph{Proc. IEEE Conf. Comput. Vis. Pattern Recognit.}, 2022, pp. 2949--2958.

\bibitem{qi2019stagnet}
M.~Qi, Y.~Wang, J.~Qin, A.~Li, J.~Luo, and L.~V. Gool, ``Stagnet: An attentive semantic rnn for group activity and individual action recognition,'' \emph{IEEE Trans. Circuits Syst. Video Technol.}, vol.~30, no.~2, pp. 549--565, 2019.

\bibitem{zhou2023hierarchical}
K.~Zhou, Y.~Ma, H.~P.~H. Shum, and X.~Liang, ``Hierarchical graph convolutional networks for action quality assessment,'' \emph{IEEE Trans. Circuits Syst. Video Technol.}, vol.~33, no.~12, pp. 7749--7763, 2023.

\bibitem{qi2020stc}
M.~Qi, Y.~Wang, A.~Li, and J.~Luo, ``Stc-gan: Spatio-temporally coupled generative adversarial networks for predictive scene parsing,'' \emph{IEEE Trans. Image Process.}, vol.~29, pp. 5420--5430, 2020.

\bibitem{gedamu2023fine}
K.~Gedamu, Y.~Ji, Y.~Yang, J.~Shao, and H.~T. Shen, ``Fine-grained spatio-temporal parsing network for action quality assessment,'' \emph{IEEE Trans. Image Process.}, vol.~32, pp. 6386--6400, 2023.

\bibitem{an2024multi}
Q.~An, M.~Qi, and H.~Ma, ``Multi-stage contrastive regression for action quality assessment,'' in \emph{Proc. IEEE Int. Conf. Acoust., Speech, Signal Process.}, 2024, pp. 4110--4114.

\bibitem{xu2024fineparser}
J.~Xu, S.~Yin, G.~Zhao, Z.~Wang, and Y.~Peng, ``Fineparser: A fine-grained spatio-temporal action parser for human-centric action quality assessment,'' in \emph{Proc. IEEE Conf. Comput. Vis. Pattern Recognit.}, 2024, pp. 14\,628--14\,637.

\bibitem{zhang2023learning}
H.~Zhang, Y.~Yuan, V.~Makoviychuk, Y.~Guo, S.~Fidler, X.~B. Peng, and K.~Fatahalian, ``Learning physically simulated tennis skills from broadcast videos,'' \emph{ACM Trans. Graph.}, vol.~42, no.~4, pp. 1--14, 2023.

\bibitem{li2022pairwise}
M.~Li, H.~Zhang, Q.~Lei, Z.~Fan, J.~Liu, and J.~Du, ``Pairwise contrastive learning network for action quality assessment,'' in \emph{Proc. Eur. Conf. Comput. Vis.}, 2022, pp. 457--473.

\bibitem{nekoui2020falcons}
M.~Nekoui, F.~O.~T. Cruz, and L.~Cheng, ``Falcons: Fast learner-grader for contorted poses in sports,'' in \emph{Proc. IEEE Conf. Comput. Vis. Pattern Recognit. Workshops}, 2020, pp. 900--901.

\bibitem{qi2019attentive}
M.~Qi, W.~Li, Z.~Yang, Y.~Wang, and J.~Luo, ``Attentive relational networks for mapping images to scene graphs,'' in \emph{Proc. IEEE Conf. Comput. Vis. Pattern Recognit.}, 2019, pp. 3957--3966.

\bibitem{lee2023hierarchically}
J.~Lee, M.~Lee, D.~Lee, and S.~Lee, ``Hierarchically decomposed graph convolutional networks for skeleton-based action recognition,'' in \emph{Proc. IEEE Int. Conf. Comput. Vis.}, 2023, pp. 10\,444--10\,453.

\bibitem{lv2024disentangled}
C.~Lv, S.~Zhang, Y.~Tian, M.~Qi, and H.~Ma, ``Disentangled counterfactual learning for physical audiovisual commonsense reasoning,'' in \emph{Proc. Adv. Neural Inf. Process. Syst.}, vol.~36, 2024, pp. 1--13.

\bibitem{nekoui2021eagle}
M.~Nekoui, F.~O.~T. Cruz, and L.~Cheng, ``Eagle-eye: Extreme-pose action grader using detail bird's-eye view,'' in \emph{Proc. IEEE Winter Conf. Appl. Comput. Vis.}, 2021, pp. 394--402.

\bibitem{zhou2024comprehensive}
K.~Zhou, R.~Cai, L.~Wang, H.~P. Shum, and X.~Liang, ``A comprehensive survey of action quality assessment: Method and benchmark,'' \emph{arXiv preprint arXiv:2412.11149}, 2024.

\bibitem{yin2025decade}
H.~Yin, P.~Parmar, D.~Xu, Y.~Zhang, T.~Zheng, and W.~Fu, ``A decade of action quality assessment: Largest systematic survey of trends, challenges, and future directions,'' \emph{arXiv preprint arXiv:2502.02817}, 2025.

\bibitem{zeng2020hybrid}
L.~A. Zeng, F.~T. Hong, W.~S. Zheng, Q.~Z. Yu, W.~Zeng, Y.~W. Wang, and J.~H. Lai, ``Hybrid dynamic-static context-aware attention network for action assessment in long videos,'' in \emph{Proc. 28th ACM Int. Conf. Multimedia}, 2020, pp. 2526--2534.

\bibitem{li2019manipulation}
Z.~Li, Y.~Huang, M.~Cai, and Y.~Sato, ``Manipulation-skill assessment from videos with spatial attention network,'' in \emph{Proc. IEEE Int. Conf. Comput. Vis. Workshops}, 2019, pp. 4385--4395.

\bibitem{yun2025semi}
W.~Yun, M.~Qi, F.~Peng, and H.~Ma, ``Semi-supervised teacher-reference-student architecture for action quality assessment,'' in \emph{Proc. Eur. Conf. Comput. Vis.}, 2025, pp. 161--178.

\bibitem{qi2021semantics}
M.~Qi, J.~Qin, Y.~Yang, Y.~Wang, and J.~Luo, ``Semantics-aware spatial-temporal binaries for cross-modal video retrieval,'' \emph{IEEE Trans. Image Process.}, vol.~30, pp. 2989--3004, 2021.

\bibitem{doughty2019pros}
H.~Doughty, W.~Mayol-Cuevas, and D.~Damen, ``The pros and cons: Rank-aware temporal attention for skill determination in long videos,'' in \emph{Proc. IEEE Conf. Comput. Vis. Pattern Recognit.}, 2019, pp. 7862--7871.

\bibitem{pan2021adaptive}
J.~Pan, J.~Gao, and W.~Zheng, ``Adaptive action assessment,'' \emph{IEEE Trans. Pattern Anal. Mach. Intell.}, vol.~44, no.~12, pp. 8779--8795, 2021.

\bibitem{xu2024procedure}
J.~Xu, Y.~Rao, J.~Zhou \emph{et~al.}, ``Procedure-aware action quality assessment: Datasets and performance evaluation,'' \emph{Int. J. Comput. Vis.}, vol. 132, pp. 6069--6090, 2024.

\bibitem{10946879}
J.~Xu, S.~Yin, and Y.~Peng, ``Human-centric fine-grained action quality assessment,'' \emph{IEEE Transactions on Pattern Analysis and Machine Intelligence}, pp. 1--13, 2025.

\bibitem{10706814}
K.~Gedamu, Y.~Ji, Y.~Yang, J.~Shao, and H.~T. Shen, ``Self-supervised sub-action parsing network for semi-supervised action quality assessment,'' \emph{IEEE Trans. Pattern Anal. Mach. Intell.}, vol.~33, pp. 6057--6070, 2024.

\bibitem{liu2024vision}
J.~Liu, H.~Wang, K.~Stawarz, S.~Li, Y.~Fu, and H.~Liu, ``Vision-based human action quality assessment: A systematic review,'' \emph{Expert Syst. Appl.}, p. 125642, 2024.

\bibitem{Tran2015Learning}
D.~Tran, L.~Bourdev, R.~Fergus, L.~Torresani, and M.~Paluri, ``Learning spatiotemporal features with 3d convolutional networks,'' in \emph{Proc. IEEE Int. Conf. Comput. Vis.}, 2015, pp. 4489--4497.

\bibitem{carreira2017quo}
J.~Carreira and A.~Zisserman, ``Quo vadis, action recognition? a new model and the kinetics dataset,'' in \emph{Proc. IEEE Conf. Comput. Vis. Pattern Recognit.}, 2017, pp. 6299--6308.

\bibitem{parmar2017learning}
P.~Parmar and B.~T. Morris, ``Learning to score olympic events,'' in \emph{Proc. IEEE Conf. Comput. Vis. Pattern Recognit. Workshops}, 2017, pp. 20--28.

\bibitem{shi2015convolutional}
X.~Shi, Z.~Chen, H.~Wang, D.~Yeung, W.~Wong, and W.~Woo, ``Convolutional lstm network: A machine learning approach for precipitation nowcasting,'' in \emph{Proc. Adv. Neural Inf. Process. Syst.}, 2015, pp. 802--810.

\bibitem{wang2021tsa}
S.~Wang, D.~Yang, P.~Zhai, C.~Chen, and L.~Zhang, ``Tsa-net: Tube self-attention network for action quality assessment,'' in \emph{Proc. 29th ACM Int. Conf. Multimedia}, 2021, pp. 4902--4910.

\bibitem{ahmadi2009towards}
A.~Ahmadi, D.~Rowlands, and D.~A. James, ``Towards a wearable device for skill assessment and skill acquisition of a tennis player during the first serve,'' \emph{Sports Technol.}, vol.~2, no. 3-4, pp. 129--136, 2009.

\bibitem{bourgain2018effect}
M.~Bourgain, S.~Hybois, P.~Thoreux, O.~Rouillon, P.~Rouch, and C.~Sauret, ``Effect of shoulder model complexity in upper-body kinematics analysis of the golf swing,'' \emph{J. Biomech.}, vol.~75, pp. 154--158, 2018.

\bibitem{okamoto2024hierarchical}
L.~Okamoto and P.~Parmar, ``Hierarchical neurosymbolic approach for comprehensive and explainable action quality assessment,'' in \emph{Proc. IEEE Conf. Comput. Vis. Pattern Recognit.}, 2024, pp. 3204--3213.

\bibitem{he2016deep}
K.~He, X.~Zhang, S.~Ren, and J.~Sun, ``Deep residual learning for image recognition,'' in \emph{Proc. IEEE Conf. Comput. Vis. Pattern Recognit.}, 2016, pp. 770--778.

\bibitem{dgcnn}
Y.~Wang, Y.~Sun, Z.~Liu, S.~E. Sarma, M.~M. Bronstein, and J.~M. Solomon, ``Dynamic graph cnn for learning on point clouds,'' \emph{ACM Trans. Graph.}, 2019.

\bibitem{cho2014learning}
K.~Cho, B.~V. Merriënboer, C.~Gulcehre, D.~Bahdanau, F.~Bougares, H.~Schwenk, and Y.~Bengio, ``Learning phrase representations using rnn encoder-decoder for statistical machine translation,'' \emph{arXiv:1406.1078}, 2014.

\bibitem{wu2019detectron2}
Y.~Wu, A.~Kirillov, F.~Massa, W.-Y. Lo, and R.~Girshick, ``Detectron2,'' \url{https://github.com/facebookresearch/detectron2}, 2019.

\bibitem{wang2020deep}
J.~Wang, K.~Sun, T.~Cheng, B.~Jiang, C.~Deng, Y.~Zhao, D.~Liu, Y.~Mu, M.~Tan, X.~Wang \emph{et~al.}, ``Deep high-resolution representation learning for visual recognition,'' \emph{IEEE Trans. Pattern Anal. Mach. Intell.}, vol.~43, no.~10, pp. 3349--3364, 2020.

\bibitem{russell2008labelme}
B.~C. Russell, A.~Torralba, K.~P. Murphy, and W.~T. Freeman, ``Labelme: A database and web-based tool for image annotation,'' \emph{Int. J. Comput. Vis.}, vol.~77, pp. 157--173, 2008.

\bibitem{andriluka20142d}
M.~Andriluka, L.~Pishchulin, P.~Gehler, and B.~Schiele, ``2d human pose estimation: New benchmark and state of the art analysis,'' in \emph{Proc. IEEE Conf. Comput. Vis. Pattern Recognit.}, 2014, pp. 3686--3693.

\bibitem{dadashzadeh2023pecopparameterefficientcontinual}
A.~Dadashzadeh, S.~Duan, A.~Whone, and M.~Mirmehdi, ``Pecop: Parameter efficient continual pretraining for action quality assessment,'' in \emph{Proc. IEEE Winter Conf. Appl. Comput. Vis.}, 2024, pp. 42--52.

\bibitem{ke2024two}
X.~Ke, H.~Xu, X.~Lin, and W.~Guo, ``Two-path target-aware contrastive regression for action quality assessment,'' \emph{Information Sciences}, vol. 664, p. 120347, 2024.

\bibitem{oord2018representation}
A.~van~den Oord, Y.~Li, and O.~Vinyals, ``Representation learning with contrastive predictive coding,'' 2018, arXiv:1807.03748.

\bibitem{schultz2003learning}
M.~Schultz and T.~Joachims, ``Learning a distance metric from relative comparisons,'' \emph{Adv. Neural Inf. Process. Syst.}, vol.~16, 2003.

\bibitem{bai2018empirical}
S.~Bai, J.~Z. Kolter, and V.~Koltun, ``An empirical evaluation of generic convolutional and recurrent networks for sequence modeling,'' 2018, arXiv:1803.01271.

\bibitem{yi2021asformer}
F.~Yi, H.~Wen, and T.~Jiang, ``Asformer: Transformer for action segmentation,'' in \emph{Proc. Brit. Mach. Vis. Conf.}, 2021, pp. 1--15.

\bibitem{selvaraju2017grad}
R.~R. Selvaraju, M.~Cogswell, A.~Das, R.~Vedantam, D.~Parikh, and D.~Batra, ``Grad-cam: Visual explanations from deep networks via gradient-based localization,'' in \emph{Proc. IEEE Int. Conf. Comput. Vis.}, 2017, pp. 618--626.

\bibitem{tsne}
L.~van~der Maaten and G.~Hinton, ``Visualizing data using t-sne,'' \emph{Journal of Machine Learning Research}, vol.~9, no.~86, pp. 2579--2605, 2008.

\bibitem{radosavovic2020designing}
I.~Radosavovic, R.~P. Kosaraju, R.~Girshick, K.~He, and P.~Doll{\'a}r, ``Designing network design spaces,'' in \emph{Proc. IEEE Conf. Comput. Vis. Pattern Recognit.}, 2020, pp. 10\,425--10\,433.

\end{thebibliography}

\begin{IEEEbiography}[{\includegraphics[width=1in,height=1in,keepaspectratio]{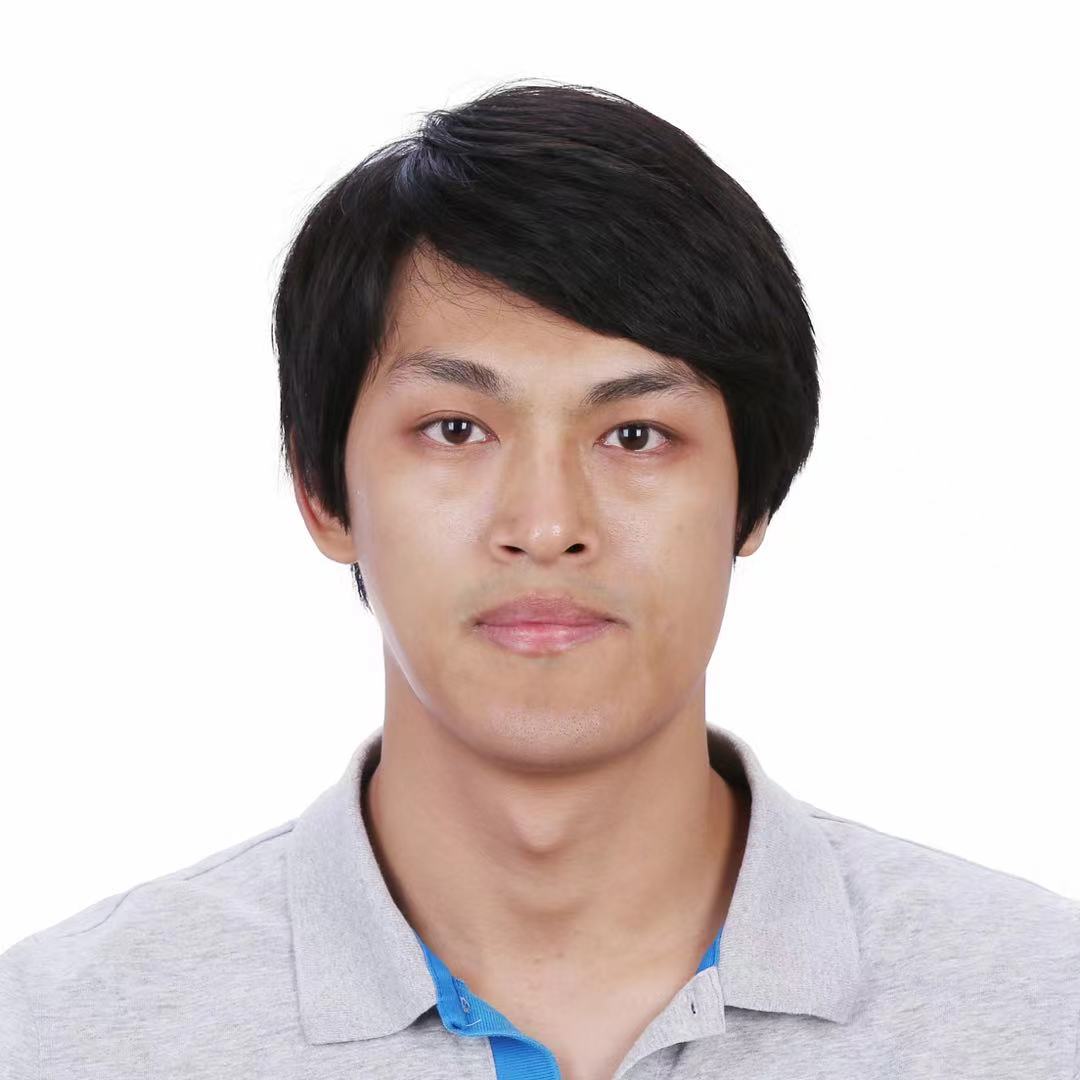}}] {Mengshi Qi} (Member, IEEE) is currently a Professor with the Beijing University of Posts and Telecommunications, Beijing, China. He received the B.S. degree from the Beijing University of Posts and Telecommunications in 2012, and the M.S. and Ph.D. degrees in computer science from Beihang University, Beijing, China, in 2014 and 2019, respectively. He was a postdoctoral researcher with the CVLAB, EPFL, Switzerland from 2019 to 2021. His research interests include machine learning and computer vision, especially scene understanding, 3D reconstruction, and multimedia analysis. He has published more than 40 papers in top journals (such as IEEE TIP, TPAMI, TMM, TCSVT, TIFS) and top conferences (such as IEEE CVPR, ICCV, ECCV, ACM Multimedia, AAAI, NeurIPS). He also has served as Area Chair of ICME 2024-2025, Senior PC Member of AAAI 2023-2025 and IJCAI 2021/2023-2025, and the Guest Editor for IEEE Transactions on Multimedia.
\end{IEEEbiography}

\begin{IEEEbiography}[{\includegraphics[width=1in,height=1in,keepaspectratio]{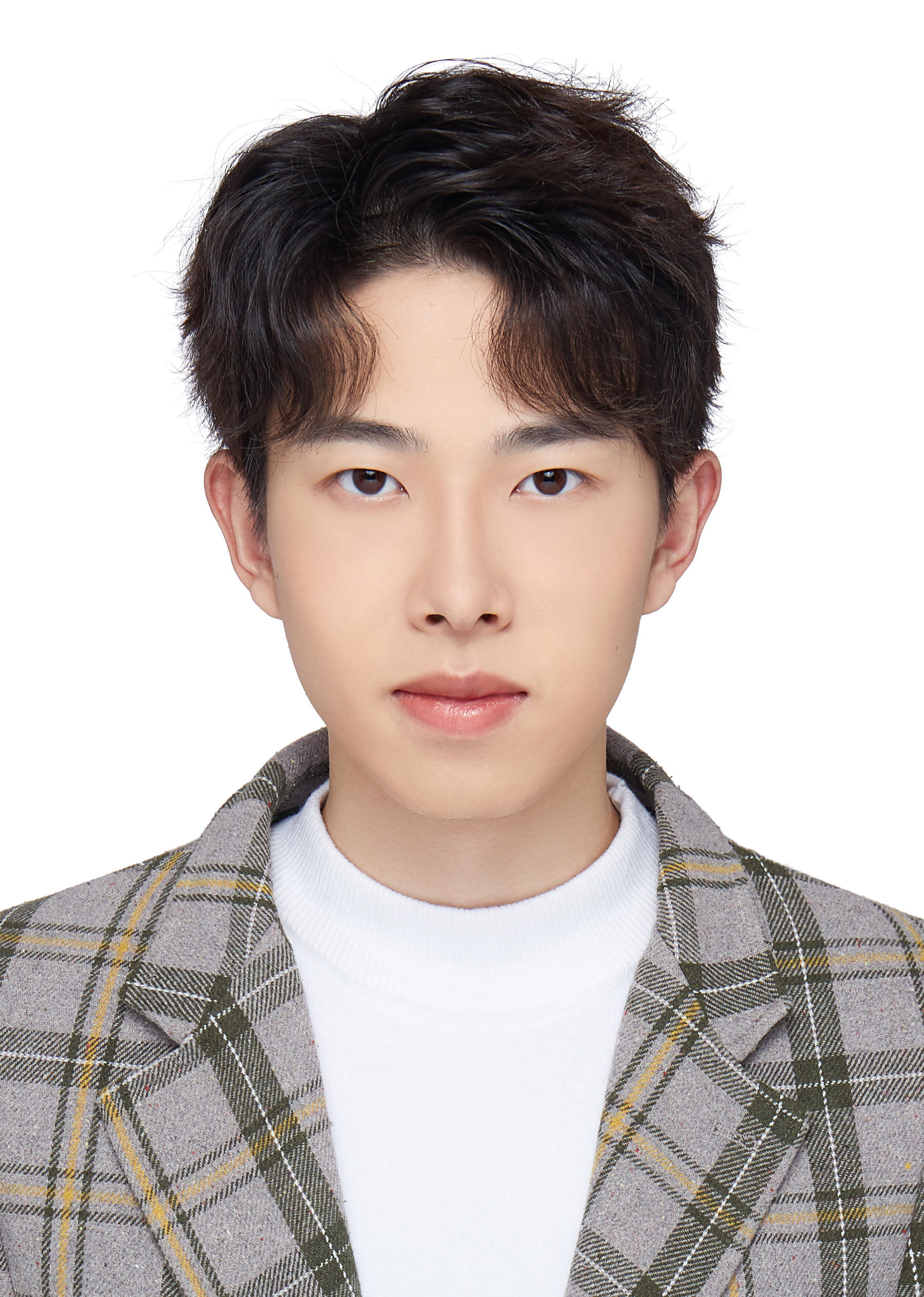}}] {Hao Ye} is currently pursuing a master's degree at the Beijing University of Posts and Telecommunications, Beijing, China. His primary research interests include multimodal learning and computer vision, with an emphasis on multimodal fusion.
\end{IEEEbiography}

\begin{IEEEbiography}[{\includegraphics[width=1in,height=1in,keepaspectratio]{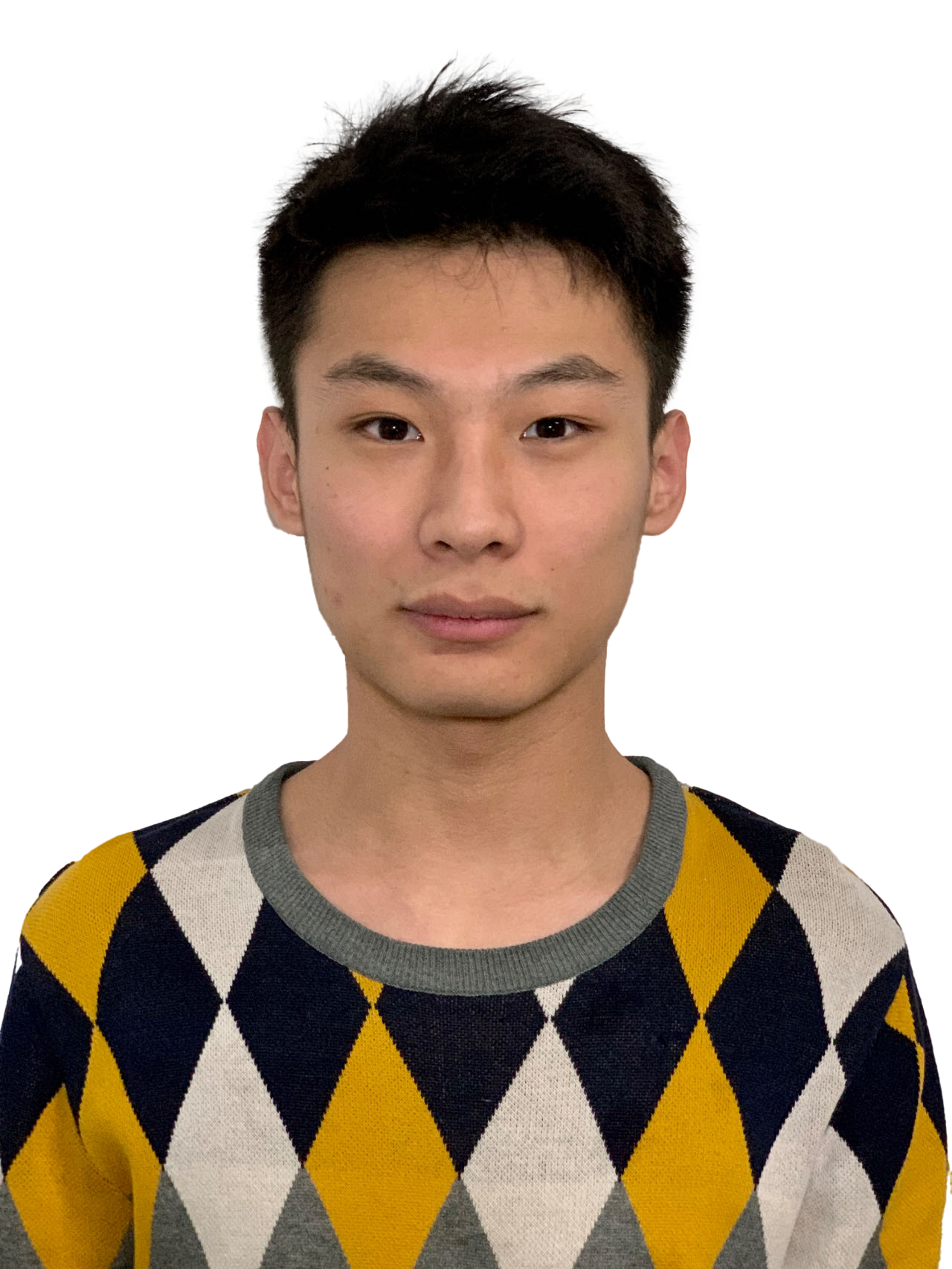}}] {Jiaxuan Peng} is currently pursuing the Master's degree at Beijing University of Posts and Telecommunications, Beijing, China. His research interests focus on computer vision and Internet of Things (IoT) applications, with an emphasis on multimodal learning in IoT.
\end{IEEEbiography}

\begin{IEEEbiography}[{\includegraphics[width=1in,height=1in,keepaspectratio]{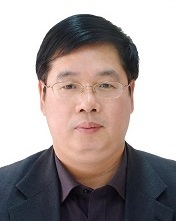}}]{Huadong Ma} (Fellow, IEEE) received the B.S. degree in mathematics from Henan Normal University, Xinxiang, China, in 1984, the M.S. degree in computer science from the Shenyang Institute of Computing Technology, Chinese Academy of Science, China, in 1990, and the Ph.D. degree in computer science from the Institute of Computing Technology, Chinese Academy of Science, Beijing, China, in 1995. He is currently a Professor of School of Computer Science, Beijing University of Posts and Telecommunications, Beijing, China. From 1999 to 2000, he held a Visiting Position with the University of Michigan, Ann Arbor, MI, USA. His current research interests include Internet of Things, sensor networks, and multimedia computing. He has authored more than 300 papers in journals, such as ACM/IEEE Transactions or conferences, such as ACM MobiCom, ACM SIGCOMM, IEEE INFOCOM, and five books. He was the recipient of the Natural Science Award of the Ministry of Education, China, in 2017, 2019 Prize Paper Award of IEEE TRANSACTIONS ON MULTIMEDIA, 2018 Best Paper Award from IEEE MULTIMEDIA, Best Paper Award in IEEE ICPADS 2010, Best Student Paper Award in IEEE ICME 2016 for his coauthored papers, and National Funds for Distinguished Young Scientists in 2009. He was/is an Editorial Board Member of the IEEE TRANSACTIONS ON MULTIMEDIA, IEEE INTERNET OF THINGS JOURNAL, ACM Transactions on Internet of Things. He is the Chair of ACM SIGMOBILE China.
\end{IEEEbiography}

\end{document}